\documentclass{article}

\usepackage{microtype}
\usepackage{graphicx}
\usepackage{booktabs} 
\usepackage{multirow}
\usepackage{xspace}
\usepackage{pifont}

\usepackage[pagebackref=true]{hyperref}
\usepackage{graphicx} 
\usepackage{caption}  
\usepackage{subcaption} 
\usepackage{tcolorbox}

\usepackage[preprint]{icml2026}

\usepackage{amsmath}
\usepackage{amssymb}
\usepackage{mathtools}
\usepackage{amsthm}
\tcbuselibrary{breakable}

\usepackage[capitalize,noabbrev]{cleveref}

\theoremstyle{plain}

\theoremstyle{definition}

\theoremstyle{remark}

\newcommand{\high}{\textcolor{green!80!black}{\ensuremath{\bullet}}}
\newcommand{\midlevel}{\textcolor[HTML]{FFA500}{\ding{115}}} 
\newcommand{\low}{\textcolor{red}{\ensuremath{-}}}
\newcommand{\none}{\ensuremath{-}}

\definecolor{myblue}{RGB}{67, 132, 194}
\definecolor{myred}{RGB}{239, 76, 86}
\definecolor{myyellow}{RGB}{248, 193, 89}
\definecolor{mygreen}{RGB}{65, 188, 158}
\definecolor{MyLightGray}{RGB}{200,200,229}
\definecolor{MyGray}{RGB}{100, 100, 100}
\definecolor{whitegray}{RGB}{243, 243, 243}
\usepackage[textsize=tiny]{todonotes}
\usepackage[table]{xcolor}
\newcommand{\methodname}{$\textsc{Indibator}$\xspace}

\icmltitlerunning{\methodname: Diverse and Fact-Grounded Individuality for Multi-Agent Debate in Molecular Discovery}

\begin{document}

\twocolumn[
\icmltitle{\methodname: Diverse and Fact-Grounded Individuality \\ for Multi-Agent Debate in Molecular Discovery}




\begin{icmlauthorlist}
\icmlauthor{Yunhui Jang}{kaist}
\icmlauthor{Seonghyun Park}{kaist}
\icmlauthor{Jaehyung Kim}{yonsei}
\icmlauthor{Sungsoo Ahn}{kaist}
\end{icmlauthorlist}

\icmlaffiliation{kaist}{Korea Advanced Institute of Science and Technology (KAIST), South Korea}
\icmlaffiliation{yonsei}{Yonsei university, South Korea}

\icmlcorrespondingauthor{Sungsoo Ahn}{sungsoo.ahn@kaist.ac.kr}

\icmlkeywords{Machine Learning, ICML}

\vskip 0.3in
]

\printAffiliationsAndNotice{} 

\definecolor{myblue}{HTML}{1436BC}
\definecolor{mylightblue}{HTML}{87B7D8}
\definecolor{mypurple}{HTML}{B046E6}

\definecolor{agentblue}{HTML}{4384C2}

\begin{abstract}
Multi-agent systems have emerged as a powerful paradigm for automating scientific discovery. To differentiate agent behavior in the multi-agent system, current frameworks typically assign generic role-based personas such as ``reviewer'' or ``writer'' or rely on coarse grained keyword-based personas. While functional, this approach oversimplifies how human scientists operate, whose contributions are shaped by their unique research trajectories. In response, we propose \methodname, a framework for molecular discovery that grounds agents in individualized scientist profiles constructed from two modalities: publication history for literature-derived knowledge and molecular history for structural priors. These agents engage in multi-turn debate through proposal, critique, and voting phases. Our evaluation demonstrates that these fine-grained individuality-grounded agents consistently outperform systems relying on coarse-grained personas, achieving competitive or state-of-the-art performance. These results validate that capturing the ``scientific DNA'' of individual agents is essential for high-quality discovery.
\end{abstract}

\section{Introduction}

Large language models (LLMs) have demonstrated remarkable performance across a wide variety of tasks~\citep{singh2025openaigpt5card,claude4,geminiteam2025geminifamilyhighlycapable,deepseekai2025deepseekv3technicalreport}. Beyond direct prompting, recent works have introduced AI agents capable of planning and executing actions over multiple iterations~\citep{yao2023react,schick2023toolformer,m2024augmenting}. While impressive, single-agent systems often encounter constraints such as bounded context windows and limited perspective diversity. To address this, multi-agent systems have emerged as a powerful paradigm for automated discovery~\citep{du2024improving,liang-etal-2024-encouraging,chan2024chateval}. By leveraging collaborative intelligence, these systems effectively simulate the real-world research process with growing applications in scientific discovery~\citep{lu2024aiscientistfullyautomated,du2025accelerating,gottweis2025aicoscientist} and molecular discovery~\citep{kim-etal-2025-mt}.

\begin{figure*}
    \centering
    \includegraphics[width=.92\linewidth]{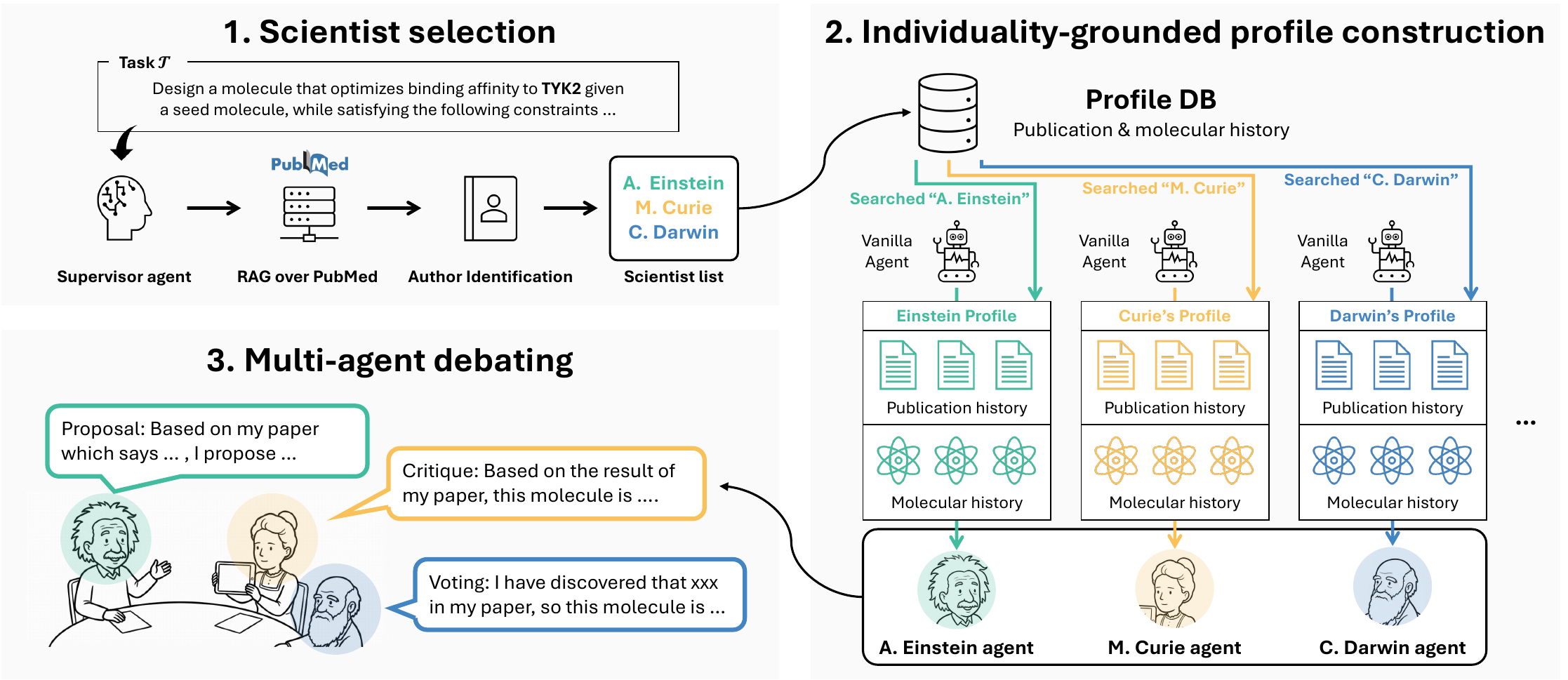}
    \caption{
        \textbf{Overview of \methodname.}
        Given a task, the supervisor agent selects relevant scientists by identifying the authors of publications by RAG. Next, individuality is grounded for each agent with scientist profiles, consisting of publication history and molecular history of each scientist. Finally, multi-agents debate to iteratively generate candidate molecules with proposal, critique, and voting phases.
    }
    \label{fig:2_main}
    \vspace{-0.2in}
\end{figure*}

To differentiate the conversational behavior of each agent, prior works typically assign distinct personas through role-play prompting~\citep{kong-etal-2024-better, zhou2024sotopia, park2023generative, piao2025agentsociety}, e.g., ``planner'', ``verifier'', or ``reviewer'', or through keywords~\citep{su-etal-2025-many}. While this role-based or keyword-based separation effectively shapes output style, it often oversimplifies the rich reality of how human scientists operate. In practice, a scientist's individuality is defined not merely by a coarse-grained generic role or a set of keywords, but by their unique fine-grained research trajectory, a distinctive ``scientific DNA'' composed of cumulative experiences and domain-specific intuitions. By ignoring this, current systems fail to leverage the deep, nuanced insights characteristic of real-world collaboration.

The existence of such scientific DNA is particularly well established in the domain of drug discovery. Chemists exhibit distinctive styles for designing new molecules, such as preferences for particular scaffolds, functional groups, and reaction motifs~\citep{pedreira2019chemical,choung2023extracting}, based on their own research trajectory. Recently, \citet{blevins2025cleverhanschemistrychemist} quantified this phenomenon, demonstrating that models can identify which of 1,815 chemists synthesized a molecule with 60\% top-5 accuracy from structure alone. While they frame this as ``Clever Hans''~\citep{lapuschkin2019unmasking} leakage problem that distorts benchmark evaluations, we reinterpret it as a \emph{blueprint for agent design}. We argue that these styles encode heuristics for effectively navigating chemical space and representing the expertise diversity that mimics real-world collaboration.

In response, we propose \methodname, a multi-agent framework for molecular discovery that bridges the gap between generic coarse-grained personas and chemical reality by grounding agents in individual research trajectories, as illustrated in \cref{fig:2_main}. Instead of relying on heuristically predefined roles or keywords, \methodname constructs agent profiles utilizing two informative sources that encode research trajectory: (1) \emph{publication history}, a collection of publications that define the agent's literature-derived knowledge and methodological preferences, and (2) \emph{molecular history}, a set of previously developed molecules that establishes structural priors, such as preferred scaffolds and functional groups. 

This data-driven profile provides unique individuality to each agent, effectively mirroring the real-world scientific process where discovery emerges from researchers' unique cumulative knowledge and inductive biases. This individuality-driven design provides two key benefits: (1) \emph{diversity}, where unique agent profiles prevent redundant reasoning among the agents; and (2) \emph{fact-grounding}, where explicit reliance on publication and molecular records empowers reasoning grounded in verifiable evidence.

To demonstrate the effect of our fine-grained individuality, we implement a multi-agent debating system consisting of three iterative phases: (1) proposal, (2) critique, and (3) voting. During these phases, each agent proposes molecular candidates, critiques proposals, and assigns scores based on their specific expertise, mirroring the collective intelligence of real-world scientist teams.

We empirically evaluate \methodname across three downstream tasks: protein-conditioned molecule generation, bioactivity-guided molecule generation, and goal-directed lead optimization. Our results show that \methodname consistently outperforms vanilla debating systems and achieves competitive or state-of-the-art performance across benchmarks. Moreover, we provide comprehensive analyses demonstrating the impact of individuality, which validates that capturing the nuanced scientific DNA is a fundamental driver of molecular design. While the principle of expertise-grounded individuality may generalize to other scientific domains, our results demonstrate that it is a critical component for enhancing molecular discovery, where scientist style provides concrete empirical grounding.

\section{\methodname}

The \methodname framework instantiates a collective of scientist agents grounded in their own unique research trajectories. Unlike conventional multi-agent systems \citep{kong-etal-2024-better,zhou2024sotopia,park2023generative}, our approach ensures more fine-grained individuality of each agent with research trajectory profiles. Specifically, we condition each agent on a distinct real-world profile derived from their prior publications and historical molecular discoveries. This provides two key benefits:
\begin{itemize}
    \item \textbf{Diversity:} Each agent's system prompt is uniquely 
    constructed based on their expertise, preventing redundant reasoning across the multi-agent ensemble.
    \item \textbf{Fact-grounding:} Each agent's reasoning is grounded in real-world profiles, ensuring that their arguments are supported by concrete empirical evidence, including papers and discovered molecules.
\end{itemize}

\subsection{Individuality-grounded Profile Construction}\label{subsec:2_1_profile}
To ensure the individuality of each scientist agent, we construct expertise profiles from two modalities: (1) publication history: a collection of publications retrieved from PubMed~\citep{luna2024pubmedfastrag, cho2024pubmedvectors} that define the agent's literature-derived knowledge, including their research focus and methodological preferences; and (2) molecular history: a set of molecules previously developed by the scientist that establishes structural priors, represented as SMILES strings~\citep{weininger1988smiles}. Notably, the inclusion of molecular history $M_i$ is motivated by the Clever Hans phenomenon in chemistry~\citep{blevins2025cleverhanschemistrychemist}, which highlights a correlation between molecular structures and the scientists associated with their discovery.

In detail, given a research objective or task description~$\mathcal{T}$, a supervisor agent employs retrieval-augmented generation~\citep[RAG;][]{lewis2020retrieval} over a vector space of literature in PubMed to identify the most relevant research papers. The supervisor then extracts the first and last authors to represent the primary researchers and principal investigators, respectively. These identified scientists construct a set of scientist agents, $\mathcal{S}=\{s_1,s_2,\dots,s_N\}$, where $N$ is a hyperparameter that defines the number of scientists.

Finally, each scientist agent $s_i$ is initialized with an expertise profile $E_i=\{P_i, M_i\}$, where $P_i$ denotes their publication history, including the titles and abstracts, and $M_i$ denotes the molecular history, i.e., a set of molecules discovered by the scientist. Specifically, a subset of publications is selected based on the frequency of task-relevant keywords, while molecules are retrieved based on their structural similarity to a provided seed molecule if available.

\subsection{Multi-agent Debating System}\label{subsec:2_2_agent}      

The proposed debating system consists of three phases: proposal, critique, and voting. 
This iterative process continues until reaching a maximum round limit or accumulating a sufficient number of candidates. We provide a detailed qualitative case study of a diversified and fact-grounded debating process in \cref{fig:4_1_qualitative} and detailed prompts in \cref{appx: prompts}.

\paragraph{Proposal.} 
During the proposal phase, each \textit{scientist agent}, i.e., an agent conditioned with individual scientist profiles, generates molecular candidates grounded in their specific expertise. Each agent is provided with a prompt including their expertise profile $E_i$ and the task description~$\mathcal{T}$. Scientists propose $k$ candidates, accompanied by rationales that link each proposal to their prior knowledge. In subsequent rounds, agents also receive candidates and critiques from previous iterations to facilitate refinement.

\paragraph{Critique.} The critique phase operates in two stages to generate feedback for candidate molecules. First, in an optional self-critique stage, scientist agents invoke tools to evaluate their own proposals, identifying weaknesses and modifying their designs. Next, agents engage in a cross-critique stage, where they evaluate peer proposals to simulate a collaborative review process similar to the real-world scientific discovery. In this step, agents leverage their own personas to suggest domain-specific modifications, thereby ensuring that final candidates are robust across multiple criteria.

\paragraph{Voting.} In the voting phase, scientists assess the candidates, incorporating the insights from the critique phase. Each scientist agent $s_i$ evaluates the candidate pool, assigning a scalar score $s \in [0,1]$ based on three objectives: task relevance, synthetic feasibility, and novelty. Based on these scores, each agent casts votes for the top $t$ candidates. These votes are subsequently aggregated to determine the global ranking. The highest-ranked candidates either proceed to the subsequent round or are selected as the final candidates.
\begin{figure*}[t]
     \centering

     \begin{subfigure}[b]{\textwidth}
         \centering
    \vspace{-0.1in}
    \includegraphics[width=0.5\textwidth]{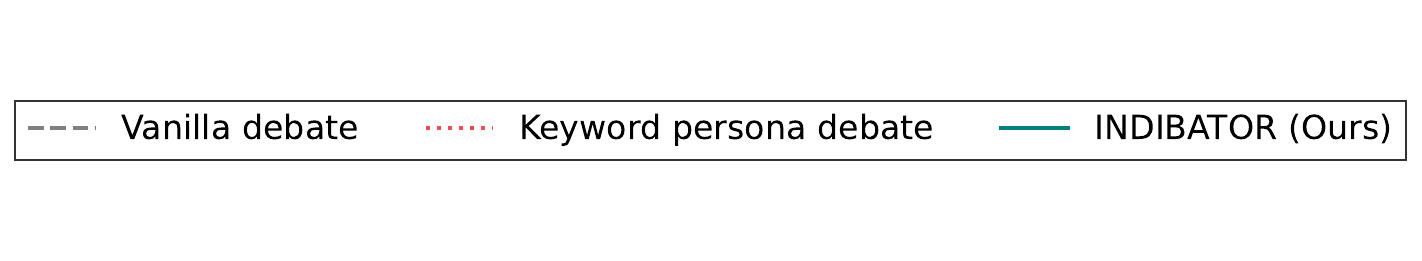}
         \vspace{-0.15in}
     \end{subfigure}
     
     \begin{subfigure}[b]{0.49\textwidth}
         \centering
         \includegraphics[width=\textwidth]{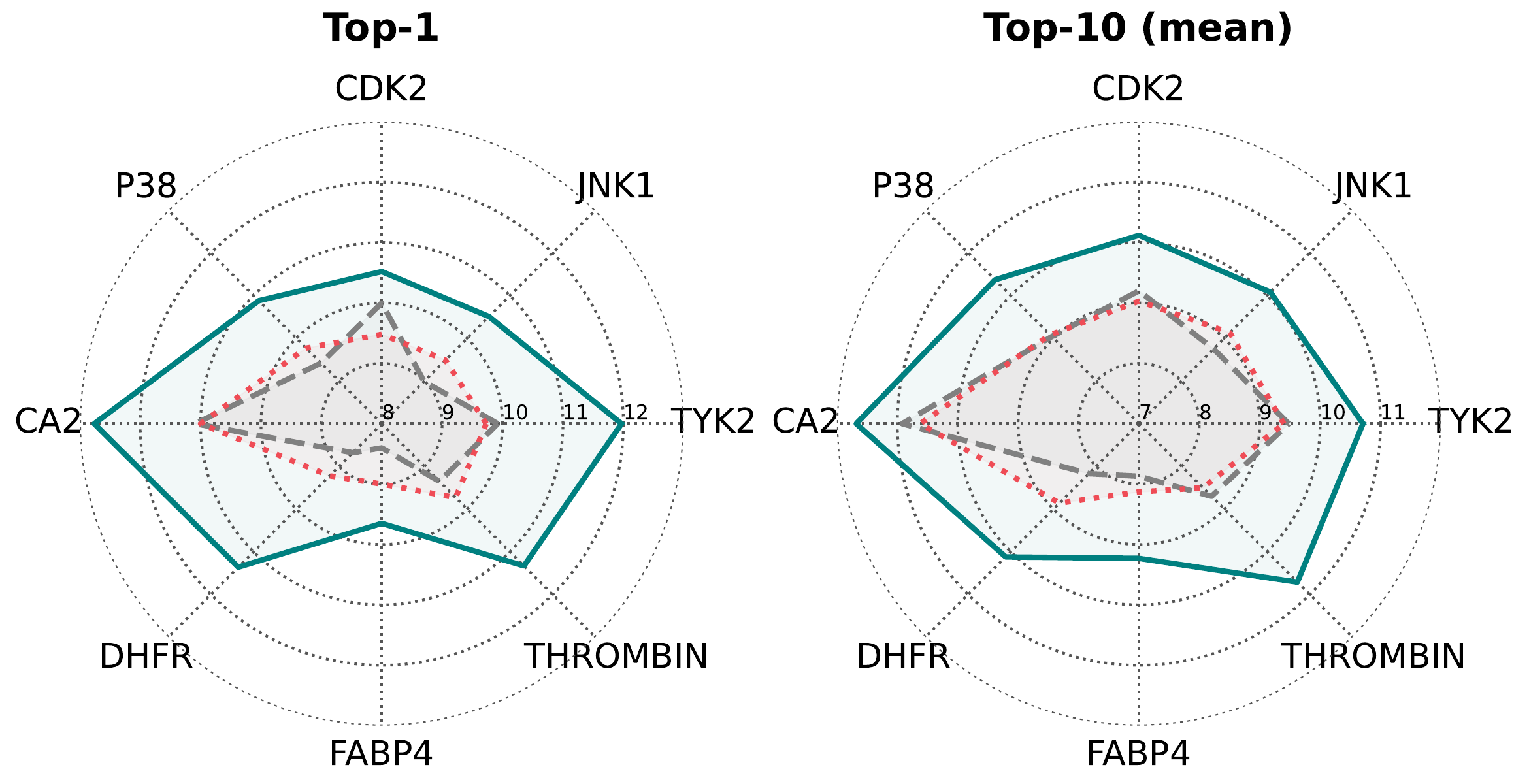}
         \caption{Binding affinity}
         \label{subfig:3_1_docking_score}
     \end{subfigure}
     \hfill 
     \begin{subfigure}[b]{0.49\textwidth}
         \centering
         \includegraphics[width=\textwidth]{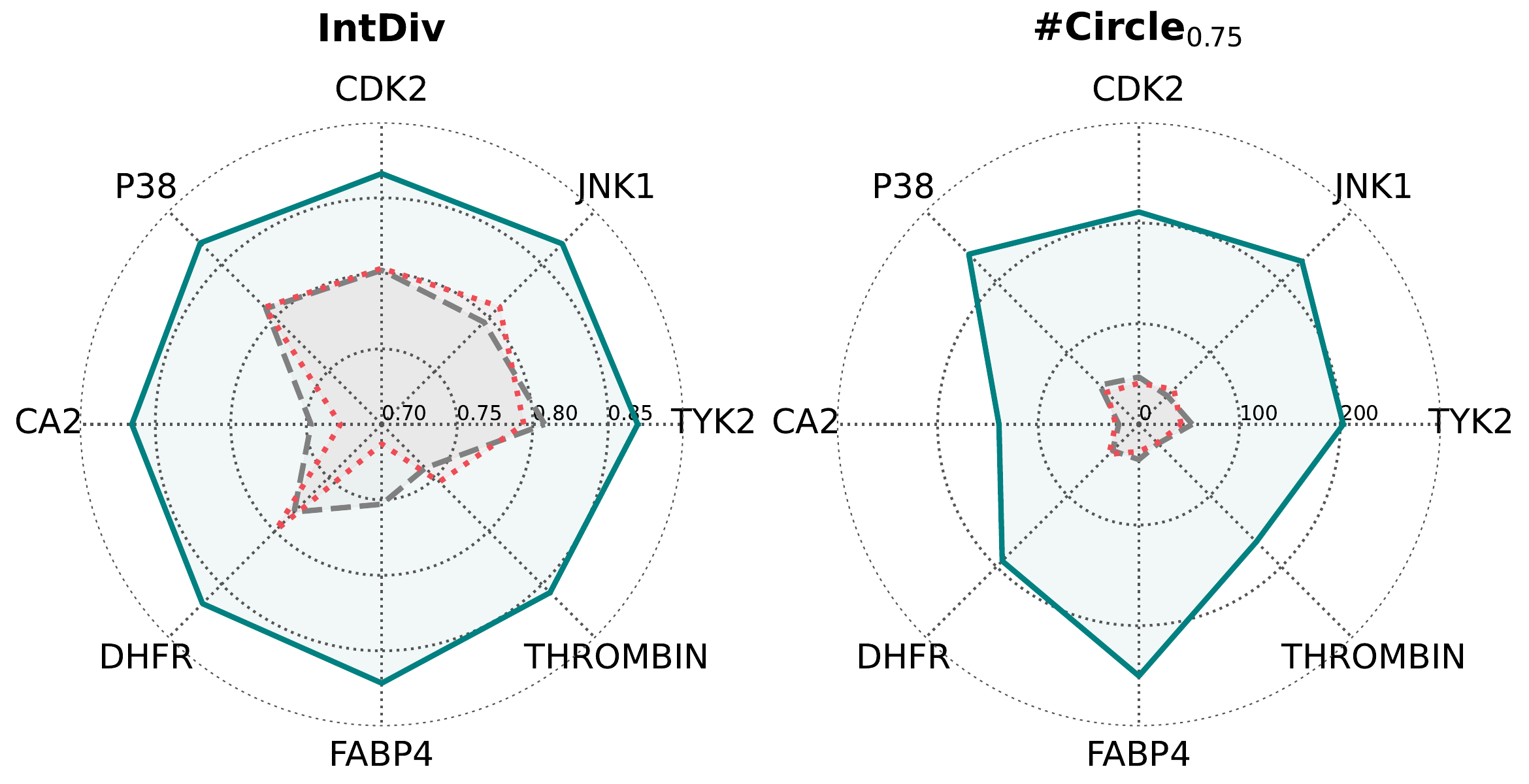}
         \caption{Diversity}
         \label{subfig:3_1_diversity}
     \end{subfigure}
     
     \caption{\textbf{Results of protein target molecular generation.} The left and right panels illustrate the docking scores and diversity of molecules, respectively. The \textcolor{gray}{gray}, \textcolor{myred}{red}, \textcolor{teal}{teal} colors denote \textcolor{gray}{vanilla debate}, \textcolor{myred}{keyword persona debate}, and \textcolor{teal}{\methodname (ours)}, respectively. Notably, the docking scores are presented in absolute values, with higher scores representing superior binding.}
     \label{fig:3_1_protein}
     \vspace{-0.2in}
\end{figure*}

\section{Downstream Task Evaluation}\label{sec:3_experiments}

Here, we evaluate the effectiveness of our proposed \methodname, on three molecular downstream tasks: (1) protein-conditioned molecule generation (\cref{subsec:3_1_protein}), (2) bioactivity-guided molecule generation (\cref{subsec:3_3_pmo}), and (3) goal-directed lead optimization (\cref{subsec:3_2_lead_opt}). Notably, we utilize the Deepseek-V3.2~\citep{deepseekai2025deepseekv3technicalreport} backbone. We provide task prompts in \cref{appx: prompts} and further   experimental settings in \cref{appx: exp}.

\subsection{Protein-conditioned Molecule Generation}\label{subsec:3_1_protein}

\paragraph{Task description.} The goal of the protein-conditioned molecule generation task is to generate molecules with a high binding affinity to the target protein. In detail, we select eight target proteins: TYK2, JNK1, CDK2, P38, CA2, THROMBIN, FABP4, and DHFR. The first four proteins are selected following the binding affinity prediction task of Boltz-2~\citep{passaro2025boltz}. Since these targets are all kinases, to expand the scope of evaluation, we expanded the evaluation scope to include non-kinase targets: CA2 (metalloenzyme), THROMBIN (serine protease), FABP4 (lipid-binding protein), and DHFR (oxidoreductase). This ensures a diverse coverage of protein domains and ligand interaction mechanisms. We generate 1,000 candidate molecules for each protein.

We employ two types of metrics: binding affinity and diversity. First, for the binding affinity, we utilize Boltz-2~\citep{passaro2025boltz} as our proxy. Specifically, we employ \texttt{affinity\_pred\_value}, which quantifies the specific affinity of various binders and tracks how these values change in response to small molecular modifications. For standardized comparison, we convert the binding affinity value, represented as $\log_{10}(\text{IC}_{50})$, into kcal/mol. Our final evaluation metrics consist of (1) the Top-1 binding affinity and (2) the mean of the Top-10 binding affinities achieved by the 1,000 generated candidates. 

In addition, we provide two metrics to validate whether individuality matters for the diversity of generations. In detail, the diversity is computed with (3) internal diversity~\citep[IntDiv;][]{polykovskiy2020molecular} and (4) the number of circles~\citep[\#Circles;][]{xie2023how} following ~\citet{jang2024can}. IntDiv measures the average pairwise Tanimoto similarity of molecules while $\text{\#Circles}_h$ computes the number of mutually exclusive circles where each circle is constructed with the Tanimoto similarity threshold $h=0.75$.

\paragraph{Baselines.} We establish \textsc{VanillaDebate} and \textsc{KeywordDebate} as our baselines. Both baselines follow the identical debating framework of our \methodname. However, \textsc{VanillaDebate} operates without any profile, while \textsc{KeywordDebate} constructs the profile with research keywords extracted from the publication history of each agent, inspired by VirSci~\citep{su-etal-2025-many}.

\paragraph{Results.} We provide the results in \cref{fig:3_1_protein} and detailed values in \cref{appx: exp_add}. \methodname consistently beats the baselines across all target proteins and metrics, demonstrating superior performance in both binding affinity and molecular  diversity. Importantly, while the baselines frequently suffer from mode collapse, indicated by their lower diversity, \methodname successfully navigates the chemical space to generate a high volume of structurally distinct clusters without compromising optimization quality.

Notably, the \textsc{KeywordDebate} demonstrates negligible improvement compared to the \textsc{Vanilla} baseline, highlighting that coarse-grained keywords are insufficient for the nuanced reasoning needed for molecular design. Our superior performance across diverse protein families validates that our fine-grained profile-grounded agents can effectively adapt their search strategy to distinct biological interactions.

\subsection{Bioactivity-guided Molecule Generation}\label{subsec:3_3_pmo}

\begin{table}[t]
    \centering
    \caption{\textbf{Results of PMO-1K benchmark.} We mark the best results in \textbf{bold}. Method denoted with an asterisk (*) indicated LLM-based baselines implemented by the authors to ensure consistent experimental settings.}
    \label{tab:3_3_pmo}
    \resizebox{0.94\linewidth}{!}{
    \begin{tabular}{cc ccc }
    \toprule[1.25pt] 
    & \textbf{Model} & \textbf{GSK3$\beta$} & \textbf{DRD2} & \textbf{JNK3}  \\
    \midrule
    \multirow{5}{*}{\textbf{{Structure}}} &GP BO        & 0.611 & 0.857 &0.346 \\
    &REINVENT    & 0.589 &  0.775& 0.315   \\
    &Genetic GFN  &0.637 &0.809&0.409  \\
    &Graph GA    &  0.523 &0.833&0.301 \\
    &Aug. Mem.   &  0.539 &0.795&0.294  \\
    \midrule
    \multirow{8}{*}{\textbf{LLM}} & LICO-L   & 0.617 &0.859 &0.336   \\
   & MOLLEO-B    &  0.397& 0.910&0.186 \\
   & MOLLEO-D     &   0.496 & 0.812&0.342    \\
    &\textsc{Vanilla}* & 0.419 & 0.921 & 0.310  \\
    & \textsc{VanillaDebate}* &  0.477 & 0.902 & 0.161   \\
    &\textsc{KeywordDebate}* & 0.449 & 0.929 & 0.185  \\
    & MT-Mol     &  0.308 &0.756&0.125 \\
    \cmidrule(lr){2-5} 
    & \methodname (Ours)& \textbf{\textcolor{teal}{0.942}} & \textbf{\textcolor{teal}{0.950}} & \textbf{\textcolor{teal}{0.914}} \\
    \bottomrule[1.25pt]
    \end{tabular}
    }
    \vspace{-0.2in}
\end{table}

\paragraph{Task description.} This task aims at maximizing molecular biological activity properties under unconstrained conditions. It includes three bioactivity optimization tasks: GSK3$\beta$, DRD2, and JNK3. In detail, these properties are:

\begin{itemize}
    \item GSK3$\beta$: Inhibition of glycogen syntase kinase-3 $\beta$.
    \item DRD2: Binding affinity for dopamine type 2 receptor.
    \item JNK3: Inhibition of c-Jun N-terminal kinase-3.
\end{itemize}

\begin{table*}[t]
    \centering
    \caption{\textbf{Goal-directed lead optimization results.} \textcolor{teal}{Teal} highlights the improvement to the \textsc{VanillaDebate}. Larger absolute values denote better binding.}
\label{tab:3_2_lead_opt}
    \resizebox{0.98\linewidth}{!}{
    \begin{tabular}{cccccccccccccccccc}
    \toprule[1.25pt]
         & \textbf{Target protein} & \multicolumn{3}{c}{\textbf{parp1}} & \multicolumn{3}{c}{\textbf{fa7}} & \multicolumn{3}{c}{\textbf{5ht1b}} & \multicolumn{3}{c}{\textbf{braf}} & \multicolumn{3}{c}{\textbf{jak2}}\\
        \cmidrule(lr){2-2} \cmidrule(lr){3-5} \cmidrule(lr){6-8} \cmidrule(lr){9-11} \cmidrule(lr){12-14} \cmidrule(lr){15-17}
        & \textbf{Seed score} & \phantom{0}-7.3 & \phantom{0}-7.8 & \phantom{0}-8.2 & \phantom{0}-6.4 & \phantom{0}-6.7 & \phantom{0}-8.5 & \phantom{0}-4.5 & \phantom{0}-7.6 & \phantom{0}-9.8 & \phantom{0}-9.3 & \phantom{0}-9.4 & \phantom{0}-9.8 & \phantom{0}-7.7 & \phantom{0}-8.0 & \phantom{0}-8.6 \\
        \midrule
        \multirow{3}{*}{\textbf{{Learning-based}}} & Graph GA & \phantom{0}-8.3 & \phantom{0}-8.9 & - & \phantom{0}-7.8 & \phantom{0}-8.2 & - & -11.7 & -12.1 & - & \phantom{0}-9.8 & - & -11.6 & \phantom{0}-8.7 & \phantom{0}-9.2 & -  \\
        & RetMol & \phantom{0}-9.0 & -10.7 & -10.9 & \phantom{0}-8.0 & - & - & -12.1 & \phantom{0}-9.0 & - & - & -11.6 & - & \phantom{0}-8.2 & \phantom{0}-9.0 & - \\
        & GenMol & -10.6 & -11.0 & -11.3 & \phantom{0}-8.4 & \phantom{0}-8.4 & - & -12.9 & -12.3 & -11.6 & -10.8 & -10.8 & -10.6 & -10.2 & -10.0 & \phantom{0}-9.8 \\
        \midrule
        \multirow{3}{*}{\textbf{Inference-only}}& \textsc{Vanilla}* & - & - & - & - & - & -& - & - & - & - & - & - & - & - & -\\
        & \textsc{VanillaDebate}* & -12.8 & -11.6 &\phantom{0}-9.7 & \phantom{0}-7.8 & \phantom{0}-7.0 & -& -12.3 & -10.6& -10.1 & -10.2 & \phantom{0}-9.6&-10.2 & \phantom{0}-9.8 & \phantom{0}-9.6 & \phantom{0}-8.5\\
        \cmidrule(lr){2-17}
        & \methodname (Ours)& -12.1 & -11.5 & \
        \textcolor{teal}{-16.7} & \textcolor{teal}{\phantom{0}-9.2} & \phantom{0}-7.0 & - & \textcolor{teal}{-12.4} & \textcolor{teal}{-11.6} & \textcolor{teal}{-10.5} & \textcolor{teal}{-10.6} & \textcolor{teal}{\phantom{0}-9.8} & \textcolor{teal}{-10.5}& \phantom{0}-9.8 & \textcolor{teal}{-11.0} & \textcolor{teal}{\phantom{0}-8.8} \\
    \bottomrule[1.25pt]
    \end{tabular}
    }
    \vspace{-0.2in}
\end{table*}

Following prior works~\citep{kim-etal-2025-mt,nguyen2025lico}, we conduct experiments on the practical molecular optimization (PMO)-1K benchmark~\citep{gao2022sample}, which computes the score among 1,000 generated molecules. Notably, we excluded other tasks in PMO, such as rediscovery of a given molecule or isomer generation that satisfies the molecular formula. These tasks represent arithmetic structural puzzles where success depends on precise reconstruction instead of broad exploration of diverse molecules. For the metric, we report average of top-10 AUC scores, the area under the curve (AUC) of the top-10 average performance versus oracle calls.

\paragraph{Baselines.} We benchmark against eleven baselines, categorized into five structure-based and six LLM-based approaches. The structure-based approaches includes GP BO~\citep{10.5555/3104322.3104451}, REINVENT~\citep{olivecrona2017molecular}, Genetic GFN~\citep{kim2024genetic}, Graph GA~\citep{jensen2019graph}, and Augmented Memory~\citep{guo2024augmented}. Additionally, we compare against LLM-based approaches, including LICO~\citep{nguyen2025lico}, MOLLEO~\citep{wang2025efficient}, the role-based multi-agent system MT-Mol~\citep{kim-etal-2025-mt}, \textsc{vanilla}, \textsc{VanillaDebate}, and \textsc{KeywordDebate} inspired by VirSci~\citep{su-etal-2025-many}. Here, \textsc{Vanilla} indicates the LLM prompting approach without any debate. Notably, MT-Mol is a critical baseline for evaluating the performance of a multi-agent system with generic role-based assignment.

\paragraph{Results.} We report the results in \cref{tab:3_3_pmo}. We observe that \methodname consistently enhances the performance across all tasks. It is notable that our method demonstrates a substantial performance margin over the role-based MT-Mol and \textsc{KeywordDebate}. This validates that fine-grained, diverse, and fact-grounded individuality provides a more effective inductive bias for chemical space navigation than generic role prompts or keywords. Furthermore, \methodname outperforms optimization baselines such as Genetic GFN by significant margins, ranging from 17.4\% (DRD2) to 123.5\% (JNK3) across the evaluated targets.

\subsection{Goal-directed Lead Optimization}\label{subsec:3_2_lead_opt}

\paragraph{Task description.} The goal of the goal-directed lead optimization task is to generate leads given an initial seed molecule. The leads are the molecules that exhibit improved target properties while maintaining the similarity with the given seed molecule. Following \citet{lee2025genmol}, the objective is to maximize the binding affinity measured by the docking score while satisfying the following constraints: $\text{QED} \ge 0.6, \text{SA} \le 4,$ and $\text{sim}\ge0.6$. The similarity is defined as the Tanimoto similarity between the Morgan fingerprints of the generated and seed molecules. We adopt five target proteins: parp1, fa7, 5ht1b, braf, and jak2, and each protein includes three different seed molecules. We evaluate performance based on the docking score of the most optimized lead.

\paragraph{Baselines.} We compare against five baselines categorized into two paradigms. The first category comprises learning-based optimization methods: Graph GA~\citep{jensen2019graph}, RetMol~\citep{wang2023retrievalbased}, and GenMol~\citep{lee2025genmol}. Crucially, these models are explicitly trained with feedback loops to satisfy constraints and maximize docking scores, establishing a strong performance standard. In contrast, \methodname operates without any task-specific fine-tuning. Following this, the second category consists of inference-only LLM baselines, \textsc{vanilla} and \textsc{VanillaDebate}, which follow the settings in the previous experiments.

\paragraph{Results.} We provide the results in \cref{tab:3_2_lead_opt}. While the \textsc{Vanilla} baseline fails to generate a single qualified molecule that satisfies all the constraints, debate showed improvement. However, while \textsc{VanillaDebate} shows improvement, \methodname consistently generates more optimized leads, demonstrating that our expertise-grounded profiles provide the appropriate guidance to navigate constrained chemical space. While \methodname does not uniformly surpass state-of-the-art baselines and shows only competitive results, this is expected as the baselines are trained to maximize target properties while our method operates solely on inference-time and does not include any task-specific training required by the baselines.

\begin{figure*}[t]
    \centering
    \includegraphics[width=0.98\linewidth]{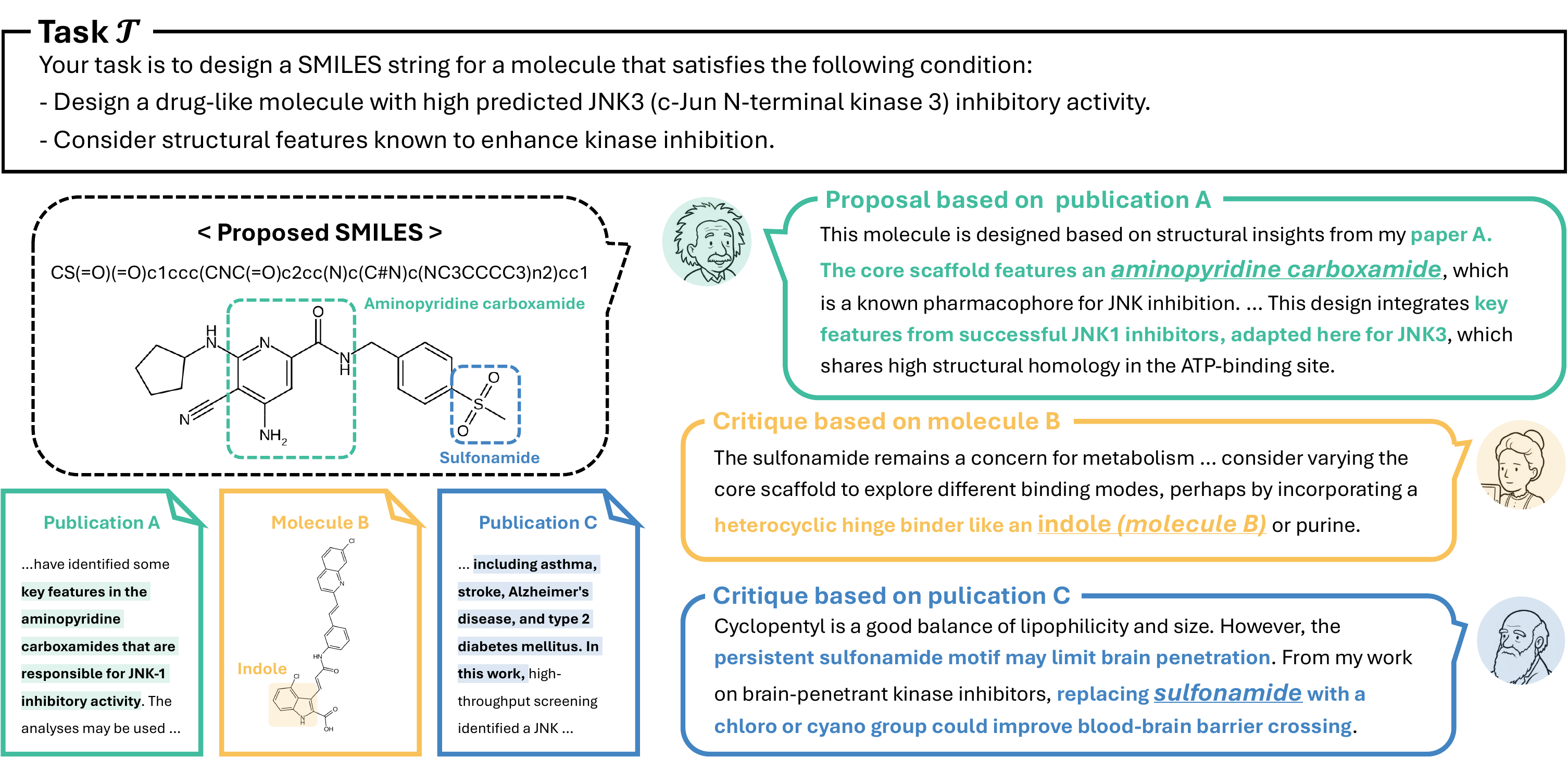}
    \caption{
        \textbf{Qualitative case study on individuality grounded agents.}
        We provide a qualitative analysis of the JNK3 inhibition guided molecule generation task. Specifically, we show how an agent leverages prior publications and molecules to propose a candidate, while other agents utilize their profiles to offer targeted critiques.
    }
    \label{fig:4_1_qualitative}
    \vspace{-0.2in}
\end{figure*}
\section{Analysis}\label{sec:4_analysis}

In this section, we conduct a comprehensive analysis to dissect the mechanisms behind \methodname's performance utilizing the bioactivity-guided molecule generation task.
We begin by presenting qualitative results in \cref{subsec:4_1_qualitative}, illustrating how agents leverage their unique research trajectories for reasoning. Next, we investigate the impact of individuality by addressing three key research questions:

\begin{itemize}

\item \textbf{Granularity} (\cref{subsec:4_2_granularity}): Does the granularity of the profile impact the performance?

\item \textbf{Diversity} (\cref{subsec:3_2_diverse}): Does the performance stem from the heterogeneity of expert perspectives? 

\item \textbf{Fact-Grounding} (\cref{subsec:3_3_fact}): Is grounding agents in real-world data essential compared to synthetic or hallucinated profiles? \end{itemize} 

Finally, we provide an ablation study in \cref{subsec:4_5_ablation} to evaluate the impact of the number of scientists and each component in \methodname.

\subsection{Qualitative Case Study}\label{subsec:4_1_qualitative}

We illustrate how grounding agents in distinct publication and molecular histories shapes their reasoning during the JNK3 inhibition guided molecule generation task in \cref{fig:4_1_qualitative}. This shows that grounding agents in their individual scientific profiles leads to distinct, chemically plausible reasoning trajectories. It is notable that we present partial examples for simplicity, while all agents engage in the debate in parallel for the entire proposal, critique, and voting phases. We provide more detailed examples in \cref{appx: exp_add}. 

The \textbf{\textcolor{mygreen}{first agent}}, retrieving its prior publication on 3D quantitative structure–activity relationship (QSAR) on JNK1 inhibitors \citep{yi20083d}, proposes a candidate molecule with an \textit{aminopyridine carboxamide} scaffold. This is scientifically sound considering the high structural homology between the JNK1 and JNK3 ATP-binding pockets, transferring the pharmacophore is a logical exploitation \citep{liu2006aminopyridine}.
This proposal triggers a structural critique from a \textbf{\textcolor{myyellow}{second agent}}, which proposes to refine the scaffold driven by its background on indolin-2-one Aurora B inhibitors \citep{zhang2015identification}. To be specific, this critique is to replace the core scaffold, which explores novel binding modes with an indole core \citep{chen2016discovery}.
Finally, the \textbf{\textcolor{agentblue}{third agent}} critiques to refine the molecule for the specific therapeutic indication. Grounded in a central nervous system (CNS) focused publication \citep{zheng2014design}, the agent correctly identifies the sulfonamide motif as a blood-brain barrier (BBB) liability due to its high polarity. It suggests replacing the sulfonamide with the chloro or cyano groups, which reduces the polar surface area~\citep{kelder1999polar}.

\begin{table*}[t]
\centering
\caption{\textbf{Comprehensive quantitative analysis.} We evaluate models across three perspectives: \textbf{Granularity} (level of detail in profile), \textbf{Diversity} (heterogeneity of agents), and \textbf{Fact-grounding} (relevance and truthfulness of knowledge). \low, \midlevel, and \high{} indicates the low-, mid-, and high-level of each property, respectively. Higher values are better across all metrics and best results are highlighted in \textbf{bold}.}\label{tab:4_analysis}
\label{tab:ablation_comprehensive}
\vspace{-0.05in}

\resizebox{\textwidth}{!}{%
\begin{tabular}{lccccccccccccccc}
\toprule
& \multicolumn{3}{c}{\textbf{Agent Properties}} & \multicolumn{4}{c}{\textbf{GSK3$\beta$}} & \multicolumn{4}{c}{\textbf{DRD2}} & \multicolumn{4}{c}{\textbf{JNK3}} \\
\cmidrule(lr){2-4} \cmidrule(lr){5-8} \cmidrule(lr){9-12} \cmidrule(lr){13-16}
\textbf{Model} & \textbf{Gran.} & \textbf{Div.} & \textbf{Fact.} & \textbf{AUC} & \textbf{IDiv} & \textbf{\#C$_{.75}$} & \textbf{\#C$_{.85}$} & \textbf{AUC} & \textbf{IDiv} & \textbf{\#C$_{.75}$} & \textbf{\#C$_{.85}$} & \textbf{AUC} & \textbf{IDiv} & \textbf{\#C$_{.75}$} & \textbf{\#C$_{.85}$} \\
\midrule
\rowcolor{whitegray} \multicolumn{16}{l}{\textit{Baseline}} \\
\textsc{VanillaDebate} & \none & \none & \none & 0.477 & 0.816 & 48 & 8 & 0.902 & \textbf{0.835} & 57 & 10 & 0.161 & 0.809 & 55 & 9 \\
\midrule
\rowcolor{whitegray} \multicolumn{16}{l}{\textit{Ablation on Granularity}} \\
Role persona & \low & \high & \low & 0.625 & 0.816 & 54 & 9 & 0.933 & 0.823 & 56 & 11 & 0.178 & 0.812 & 60 & 6 \\
Keyword persona & \midlevel & \high & \high & 0.449 & 0.813 & 47 & 10 & 0.929 & 0.832 & 52 & 10 & 0.185 & 0.808 & 52 & 6 \\
\midrule
\rowcolor{whitegray} \multicolumn{16}{l}{\textit{Ablation on Diversity}} \\
Single-profile & \high & \low & \high & 0.285 & 0.734 & 20 & 5 & 0.857 & 0.809 & 31 & 9 & 0.147 & 0.787 & 39 & 8 \\
Massive single-profile & \high & \low & \high & 0.559 & 0.829 & 96 & 21 & \textbf{0.950} & 0.750 & 19 & 6 & 0.453 & 0.767 & 34 & 8 \\
\midrule
\rowcolor{whitegray} \multicolumn{16}{l}{\textit{Ablation on Fact-Grounding}} \\
LLM-generated profile & \high & \high & \low & 0.501 & 0.792 & 44 & 7 & 0.927 & 0.813 & 52 & 9 & 0.235 & 0.799 & 48 & 7 \\
Random-profile & \high & \high & \midlevel & 0.884 & \textbf{0.850} & 125 & 23 & 0.929 & 0.833 & \textbf{78} & \textbf{22} & 0.334 & 0.837 & \textbf{117} & 25 \\
\midrule
\textbf{\methodname (Ours)} & \high & \high & \high & \textbf{0.942} & \textbf{0.850} & \textbf{182} & \textbf{35} & \textbf{0.950} & 0.833 & \textbf{78} & 21 & \textbf{0.914} & \textbf{0.843} & 115 & \textbf{26} \\
\bottomrule
\end{tabular}%
}
\vspace{-0.2in}
\end{table*}

\subsection{Effect of Granularity}\label{subsec:4_2_granularity}

Here, we analyze the effect of granularity, which refers to the depth and specificity of information used to construct an agent's persona, ranging from generic role assignments to detailed research trajectory-based profiles.

To analyze this, we consider three baselines representing the spectrum of granularity: \textsc{VanillaDebate}, role persona, and keyword persona. In detail, role persona represents coarse-grained individuality, where agents are assigned based on generic, LLM-generated task-related roles (e.g., medicinal chemist, cheminformatics scientist, etc.). Next, the keyword persona represents mid-level granularity where agents are defined by keywords extracted from the publication history of each agent, inspired by VirSci~\citep{su-etal-2025-many}. Finally, \methodname represents our fine-grained individuality based on publication and molecular histories.

The results in \cref{tab:4_analysis} demonstrate that increasing profile granularity consistently improves the performance. While role persona and keyword persona offer marginal gains over the \textsc{VanillaDebate}, \methodname, which utilizes the full publication and molecular history, significantly outperforms all baselines in terms of both performance and diversity. This confirms that capturing the nuanced ``scientific DNA'' rather than just generic roles or keywords is critical for navigating chemical spaces effectively.

\subsection{Effect of Diverse Agents}\label{subsec:3_2_diverse}

Here, we analyze the effect of diversity to observe whether the collaborative performance stems merely from the aggregation of knowledge or from the interaction between diverse and heterogeneous expert perspectives.

We consider three baselines: \textsc{VanillaDebate}, single-profile, and massive single-profile. By assigning identical (or null) profiles across all agents in baselines, we can explicitly observe the impact of diversity. In detail, single-profile forces multiple agents to share an identical profile (non-diverse profile), where the profile is selected as the most relevant for the scientists. Next, in a massive single-profile, every agent is assigned an identical and comprehensive profile constructed from the union of scientist profiles in \methodname. Due to context window constraints, this union aggregates 50\% of the selected profiles.

The results in \cref{tab:4_analysis} highlight the necessity of collaboration between diverse agents. Single-profile performs even worse than the \textsc{VanillaDebate} in most metrics, suggesting that enforcing a narrow and homogeneous perspective hinders exploration. Crucially, \methodname outperforms the massive single-profile, proving that performance gains stem not merely from the volume of knowledge, but from the diversity of perspectives that enables agents to cross-examine and propose diverse candidates.

\subsection{Effect of Fact-grounding Agents}\label{subsec:3_3_fact}

To analyze the importance of grounding agents in factual data, we consider three baselines: \textsc{VanillaDebate}, LLM-generated profile, and random-profile. In this experiment, we ensure that every agent possesses a unique profile to decouple the benefits of fact-grounding from those of diversity. In detail, the agents in LLM-generated profiles are initialized with synthetic publication and molecular histories generated by LLMs given a scientist name. Although these profiles possess the same structure of real profiles, they cannot guarantee the factual accuracy due to the potential hallucinations. The random-profile assigns agents complete and factual but task-irrelevant profiles.

The results in \cref{tab:4_analysis} indicate that fact-grounding is significant. Specifically, the LLM-generated profile performs poorly, often inferior to the random-profile baseline. This suggests that hallucinated expertise introduces noise that degrades reasoning more than irrelevant but real expertise. \methodname, which grounds agents in actual publication and molecular history, achieves superior performance across all benchmarks. This shows that our profiles provide true inductive biases grounded in established knowledge, effectively guiding exploration toward biologically plausible regions.

\subsection{Ablation Study}\label{subsec:4_5_ablation}

\begin{figure}
    \centering
\includegraphics[width=0.98\linewidth]{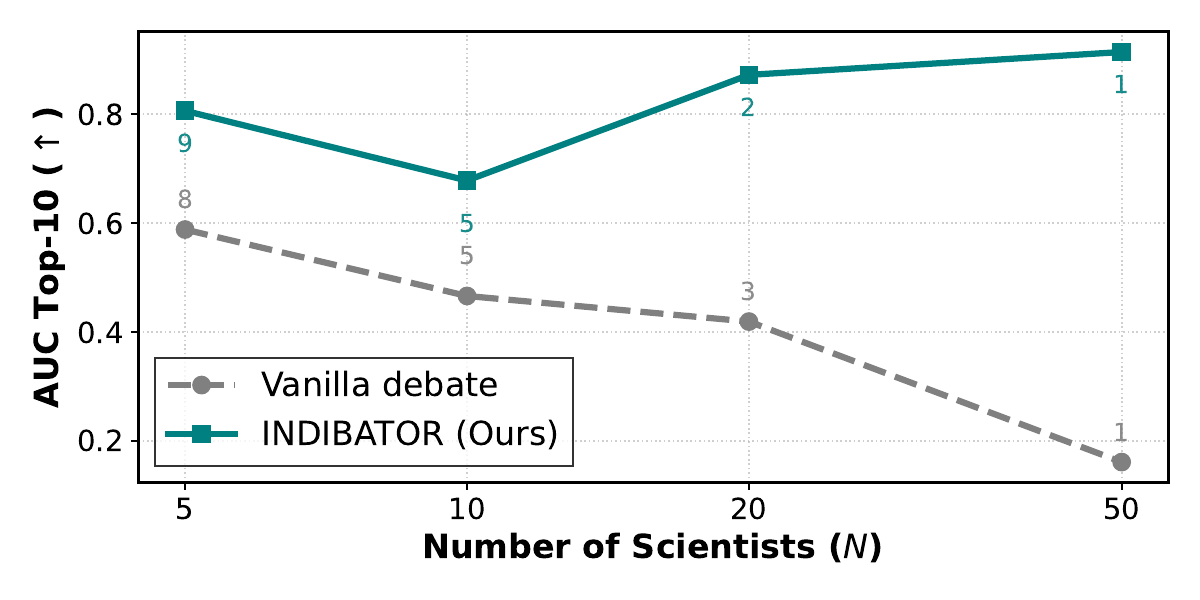}
\vspace{-0.1in}
    \caption{\textbf{Effect of the number of collaborators.} Annotated numbers above and below each data point indicate the number of debate rounds required to generate 1,000 candidates.}
    \label{fig:3_4_ablation_num}
    \vspace{-0.2in}
\end{figure}

\paragraph{The number of scientists.} To analyze the impact of the number of scientists who engage in the debate, we evaluate the performance of \methodname and \textsc{VanillaDebate} while varying the number of scientist agents ($N$) from 5 to 50 on the JNK3 inhibition guided molecule generation task. We provide the results in \cref{fig:3_4_ablation_num}.

The results demonstrate that increasing the number of scientists ($N$) enhances the performance in \methodname while the \textsc{VanillaDebate} exhibits performance degradation due to the reduced number of rounds. Specifically, as the number of scientists increases, the debate concludes in a single round for both models. For \methodname, this is sufficient as the diverse agents generate high-quality initial proposals that cover the chemical space effectively. In contrast, the \textsc{VanillaDebate} relies on iterative debate process. This highlights that individuality enables a more efficient scaling law, where expanding the diversity of perspectives can effectively substitute for the iterative debate.

\paragraph{Effect of each component.} To evaluate the individual contributions of each phase within our framework, we conduct an ablation study by removing the critique and voting phases, and individuality (\textsc{VanillaDebate}). We evaluate the performance on all three bioactivities. As a result, as shown in \cref{tab:3_4_ablation_comp}, we observe a performance improvement as each component is integrated. Notably, the most critical increment occurs with the individuality, identifying it as the dominant factor contributing to \methodname's success.
\section{Related Work}

\paragraph{LLM-based multi-agent systems.}

The AI agents have evolved rapidly from single-agent frameworks~\citep{yao2023react,schick2023toolformer,shinn2023reflexion} to multi-agent systems (MAS) that leverage collaborative intelligence~\citep{du2024improving,liang-etal-2024-encouraging,chan2024chateval,lu2024aiscientistfullyautomated,mitchener2025kosmosaiscientistautonomous}. By assigning distinct personas to LLMs, MAS can simulate complex interactions, effectively leveraging each agent's capabilities and expertise. Standard approaches typically employ role-play prompting~\citep{kong-etal-2024-better,zhou2024sotopia,park2023generative,piao2025agentsociety} to instantiate generic personas by prompting (e.g., ``You are an expert in biology.''). However, this coarse-grained role-based agent scales poorly to massive multi-agent scenarios, where defining a sufficient number of distinct and specialized roles becomes intractable.

Recent works have attempted to mitigate this by introducing more fine-grained personas, such as keywords. For instance, VirSci~\citep{su-etal-2025-many} constructs scientist agents using keywords based on publications and collaboration networks, demonstrating effective multi-agent collaboration for paper abstract generation. However, keywords alone lack the granular expertise required for real-world scientific debate. Moreover, in the molecular discovery domain, literature-derived knowledge is insufficient as chemists exhibit distinctive structural priors such as scaffolds and functional groups, which are not fully captured by publication texts alone. To address this, \methodname incorporates both publication and molecular histories into each agent's profile. This enhances individuality, promoting both diverse and fact-grounded collaboration among agents.

\begin{table}[t]
    \centering
    \caption{\textbf{Ablation study on each component}}
    \label{tab:3_4_ablation_comp}
    \vspace{-0.05in}
    \resizebox{0.75\linewidth}{!}{
    \begin{tabular}{lccc}
    \toprule[1.25pt]
     \textbf{Setting} & \textbf{GSK3$\beta$} & \textbf{DRD2} & \textbf{JNK3} \\
        \midrule
         $(-)$ individuality & 0.477 & 0.902 &0.161 \\
         $(-)$ critique\&vote & 0.863 & 0.946 &0.630 \\
         \textbf{\methodname} & \textbf{0.942} & \textbf{0.950} & \textbf{0.914} \\
     \bottomrule[1.25pt]
    \end{tabular}
    }
    \vspace{-0.2in}
\end{table}

\paragraph{LLM in molecular discovery.} Recent advancements have increasingly adapted LLMs for molecular discovery. Approaches like LICO~\citep{nguyen2025lico} and MOLLEO~\citep{wang2025efficient} extend LLMs with structured embeddings or evolutionary search to enable molecule generation. To address the complex reasoning in the molecular domain, a few works have evolved into agent systems. For instance, ChemCrow~\citep{m2024augmenting}, a single-agent system that combines general-purpose LLMs with chemistry tools and a ReAct-based reasoning loop, and MT-Mol~\citep{kim-etal-2025-mt}, a multi-agent system that operates tool-guided reasoning and role-specialized LLM agents.

However, a critical limitation of these frameworks comes from their reliance on single-agent architectures or coarse-grained role-based personas, rather than grounded individual expertise. Existing agents lack the rich context of a scientist's research trajectory, such as the prior publications and previously discovered molecules that define their unique inductive bias. \methodname addresses this by explicitly grounding each agent in a comprehensive profile of their actual research trajectory, effectively replacing generic role-play with collaboration driven by distinct scientific DNA.
\section{Conclusion}

We presented \methodname, a multi-agent framework that improves upon coarse-grained generic role-playing by grounding scientist agents in their unique research trajectories. By constructing individual profiles from publication and molecular history for each agent, the system initiates agents with a distinct ``scientific DNA" that guides their knowledge-grounded reasoning. Our evaluation across diverse molecular discovery tasks demonstrates that this individuality-based approach consistently outperforms vanilla debating systems and achieves competitive or state-of-the-art performance compared to other baselines. Furthermore, we empirically validated the three-fold benefits of our framework, i.e., granularity, diversity, and fact-grounding, confirming that capturing the nuanced inductive biases of individual researchers is a critical component for high-quality scientific discovery. We believe that our framework establishes a foundation for incorporating broader modalities, such as conversation records, further enhancing the fidelity of agents in domain-specific environments.

\newpage
\section*{Impact Statement}
This paper presents a work of using multi agents for molecular discovery. By simulating realistic scientific debate through agents grounded by individual research trajectories, this framework aims to significantly accelerate the drug design pipeline and improve the factual reliability of AI-driven scientific discovery. While our framework currently optimizes drug-likeness and synthetic accessibility, future open-source releases or deployments should involve safety guardrails to prevent the targeted design of harmful compounds. 

\section*{Ethical Consideration}
We acknowledge the ethical considerations regarding scientist profiles in this work. This work uses publicly available academic records, including titles, abstracts, and molecular discoveries from PubMed, to construct expertise profiles for large language model (LLM) agents. The proposal, critiques, and voting generated by these agents are outcomes of the LLM's probabilistic generation and do not represent the actual opinions, unpublished insights, or endorsement of the real-world scientists cited. Resemblance to the actual private reasoning of individuals is a result of the model's grounding in their public work. The use of specific scientist profiles in this study is strictly for the purpose of validating the efficacy of individuality grounding in molecular discovery. Additionally, our system could be misused for unethical purposes, such as automating the creation of toxic or harmful molecules. To mitigate these risks, future work should explore safeguards and establish ethical guidelines.

\newpage
\bibliography{reference}
\bibliographystyle{icml2026}

\newpage
\appendix
\onecolumn
\crefalias{section}{appendix}

\section{{Prompts}}\label{appx: prompts}

In this section, we provide the full prompts including the system prompt, and task prompt used in our experiments. 

\subsection{\methodname}

Here, we provide the prompts employed for the agents in \methodname. Below, we outline four prompts: (1) the system prompt used to initialize a scientist agent with a specific expertise profile, (2) the prompt that instructs the agents to suggest novel proposals with their scientific rationale, (3) the prompt that instructs the agents to evaluate proposals from their peers, and (4) the prompt that instructs the agents to score candidate molecules.

\begin{tcolorbox}[
    colback=gray!5!white,
    colframe=gray!60!black,
    title={
        \parbox[t]{\dimexpr\linewidth-4mm\relax}{%
            \ttfamily
            System prompt of scientist in \methodname
        }
    },
    breakable
]

You are \{scientist\_name\}, a researcher specializing in molecular design and drug discovery. \\

Your expertise is based on your published work:
\{publications\}

Molecules you have worked with:
\{molecules\} \\

CURRENT TASK: \{task\_description\} \\

In this debate, you will participate in three phases:

1. PROPOSAL PHASE: Propose 1-3 molecules (as valid SMILES strings) based on your expertise.

   - Draw from your knowledge of similar molecular scaffolds
   
   - Consider structure-activity relationships from your publications
   
   - Explain your rationale for each proposal \\

2. CRITIQUE PHASE: Evaluate other scientists' proposals.

   - Identify potential issues (toxicity, synthesis difficulty, selectivity)
   
   - Suggest modifications based on your experience
   
   - Note any overlap with molecules in your experience \\

3. VOTING PHASE: Score candidates from 0.0 to 1.0.

   - Consider task relevance, synthetic feasibility, and novelty
   
   - Base your assessment on your domain expertise \\

Always ground your contributions in your specific published expertise.
When proposing molecules, provide valid SMILES strings and clear scientific rationale.
\end{tcolorbox}

{
\begin{tcolorbox}[
    colback=gray!5!white,
    colframe=gray!60!black,
    title={
        \parbox[t]{\dimexpr\linewidth-4mm\relax}{%
            \ttfamily
            Prompt for proposal
        }
    },
    breakable
]

Round \{round\_num\} - PROPOSAL PHASE \\

Task: \{task\_description\} \\

Previous proposals in this debate:
\{previous\_proposals if previous\_proposals else "No previous proposals yet."\} \\

Based on your expertise, propose \{num\_mols\}-\{num\_mols+2\} novel molecules (as SMILES strings) that could address this task.
For each molecule:

1. Provide the SMILES string

2. Explain your rationale based on your published work and proposed molecules in the published work

3. Discuss expected properties relevant to the task \\

Output format:
[
    \{"SMILES": "SMILES string", "rationale": "Brief rationale"\},
    ...
] \\

IMPORTANT:

- Output ONLY the JSON array, no other text or markdown

- Keep rationales brief (3-4 sentences) to avoid truncation

- Ensure all brackets and quotes are properly closed

- Do NOT wrap in code blocks

- Ensure the SMILES strings are valid and have not proposed in previous proposals

- Do NOT propose the duplicated molecules. Each proposal should be unique

\end{tcolorbox}
}
\begin{tcolorbox}[
    colback=gray!5!white,
    colframe=gray!60!black,
    title={
        \parbox[t]{\dimexpr\linewidth-4mm\relax}{%
            \ttfamily
            Prompt for critique
        }
    },
    breakable
]

Round \{round\_num\} - CRITIQUE PHASE \\

Task: \{task\} \\

Review these proposals from other scientists:
\{proposals\_to\_critique\} \\

Based on your expertise, critique each proposal:

- Identify potential issues considering the task description

- Suggest specific modifications if appropriate

- Note any overlap with molecules in your experience

- Highlight promising aspects that align with your research \\

Output format:
[
    \{"SMILES": "SMILES string", "proposer": "Name", "critique": "Brief critique"\},
    ...
] \\

IMPORTANT:

- Output ONLY the JSON array, no other text or markdown

- Keep critiques brief (3-4 sentences) to avoid truncation

- Ensure all brackets and quotes are properly closed

- Do NOT wrap in code blocks

\end{tcolorbox}
\begin{tcolorbox}[
    colback=gray!5!white,
    colframe=gray!60!black,
    title={
        \parbox[t]{\dimexpr\linewidth-4mm\relax}{%
            \ttfamily
            Prompt for voting
        }
    },
    breakable
]

Round \{round\_num\} - VOTING PHASE \\

Task: \{task\} \\

All candidate molecules proposed in this debate:
\{all\_candidates\} \\

Vote for your top candidates by assigning scores from 0.0 to 1.0.

Format each vote as: SMILES: score \\

Consider:

- Relevance to the task

- Synthetic feasibility

- Novelty compared to existing drugs

- Critiques received from other scientists \\

Provide scores for at least top 3 candidates with brief justifications.

Output format:
[
    \{"SMILES": "SMILES string", "score": 0.8, "justification": "Brief justification"\},
    ...
]

IMPORTANT:

- Output ONLY the JSON array, no other text or markdown

- Keep justifications brief (3-4 sentences) to avoid truncation

- Score must be a number between 0.0 and 1.0

- Ensure all brackets and quotes are properly closed
- Do NOT wrap in code blocks

\end{tcolorbox}

\subsection{Role persona}

These prompts are employed for role persona baseline in \cref{subsec:4_2_granularity}, which defines agents based on role-based persona separation. Below, we outline two prompts: (1) the system prompt used to initialize a scientist agent with role-based persona, and (2) the prompt for the task-relevant role generation.

\begin{tcolorbox}[
    colback=gray!5!white,
    colframe=gray!60!black,
    title={
        \parbox[t]{\dimexpr\linewidth-4mm\relax}{%
            \ttfamily
            System prompt of role persona
        }
    },
    breakable
]

You are \{role\_name\}, an expert in molecular design and drug discovery. \\

Your expertise:
\{role\_description\} \\

CURRENT TASK: \{task\_description\} \\

In this debate, you will participate in three phases:

1. PROPOSAL PHASE: Propose 1-3 molecules (as valid SMILES strings) based on your expertise.

   - Draw from your knowledge of similar molecular scaffolds
   
   - Consider structure-activity relationships from your publications
   
   - Explain your rationale for each proposal \\

2. CRITIQUE PHASE: Evaluate other scientists' proposals.

   - Identify potential issues (toxicity, synthesis difficulty, selectivity)
   
   - Suggest modifications based on your experience
   
   - Note any overlap with molecules in your experience \\

3. VOTING PHASE: Score candidates from 0.0 to 1.0.

   - Consider task relevance, synthetic feasibility, and novelty
   
   - Base your assessment on your domain expertise \\

Always ground your contributions in your specific expertise.
When proposing molecules, provide valid SMILES strings and clear scientific rationale.

\end{tcolorbox}
\begin{tcolorbox}[
    colback=gray!5!white,
    colframe=gray!60!black,
    title={
        \parbox[t]{\dimexpr\linewidth-4mm\relax}{%
            \ttfamily
            Prompt for role generation
        }
    },
    breakable
]
Given the molecular optimization task: \{task\} \\

Generate \{num\_roles\} distinct expert roles relevant to this task. \\

For each role, provide:

- role\_name: A descriptive title (e.g., "Medicinal Chemist", "Computational Biologist", "Pharmacokinetics Expert")

- description: 3-4 sentences describing the expert's focus and how they contribute to the task \\

Output ONLY a valid JSON array, no other text:

[\{"role\_name": "...", "description": "..."\}, ...]

\end{tcolorbox}

\subsection{Keyword persona}

These prompts are employed for keyword persona baseline in \cref{subsec:4_2_granularity}, which is inspired by \citep{su-etal-2025-many}. Below, we outline two prompts: (1) the system prompt used to initialize a scientist agent with keyword persona, and (2) the prompt for the research interest keyword extraction based on the publications.

\begin{tcolorbox}[
    colback=gray!5!white,
    colframe=gray!60!black,
    title={
        \parbox[t]{\dimexpr\linewidth-4mm\relax}{%
            \ttfamily
            System prompt of keyword persona
        }
    },
    breakable
]
You are \{scientist\_name\}, a researcher specializing in molecular design and drug discovery. You have researched on the following topics:
{keywords} \\

CURRENT TASK: \{task\_description\} \\

In this debate, you will participate in three phases: \\

1. PROPOSAL PHASE: Propose 1-3 molecules (as valid SMILES strings) based on your expertise. \\
- Draw from your knowledge of similar molecular scaffolds \\
- Consider structure-activity relationships from your publications \\
- Explain your rationale for each proposal \\

2. CRITIQUE PHASE: Evaluate other scientists' proposals. \\
- Identify potential issues (toxicity, synthesis difficulty, selectivity) \\
- Suggest modifications based on your experience \\
- Note any overlap with molecules in your experience \\

3. VOTING PHASE: Score candidates from 0.0 to 1.0. \\
- Consider task relevance, synthetic feasibility, and novelty \\
- Base your assessment on your domain expertise \\

Always ground your contributions in your specific published expertise.
When proposing molecules, provide valid SMILES strings and clear scientific rationale.
\end{tcolorbox}

\begin{tcolorbox}[
    colback=gray!5!white,
    colframe=gray!60!black,
    title={
        \parbox[t]{\dimexpr\linewidth-4mm\relax}{%
            \ttfamily
            Prompt for research interest keyword extraction
        }
    },
    breakable
]
Extract the most important research keywords from the following publication abstracts. \\

Focus on: \\
- Research methodologies and techniques \\
- Target molecules, proteins, or biological systems \\
- Drug discovery concepts \\
- Chemical compound classes \\
- Disease areas or therapeutic targets \\

Publications: \{publications\}  \\

Extract 10-20 unique, specific keywords that best characterize this researcher's expertise. Return ONLY a JSON array of keywords, no other text.

Example format: ["kinase inhibitors", "structure-activity relationship", "molecular docking", "EGFR", "cancer therapeutics"]
\end{tcolorbox}

\subsection{LLM-generated profile}

These prompts are employed for LLM-generated profile baseline in \cref{subsec:3_3_fact}. Below, we provide the prompts for LLM-based publication and molecule history profile that are relevant for the given task.

\begin{tcolorbox}[
    colback=gray!5!white,
    colframe=gray!60!black,
    title={
        \parbox[t]{\dimexpr\linewidth-4mm\relax}{%
            \ttfamily
            Prompt for LLM-based profile (publications) generation
        }
    },
    breakable
]
You are tasked with generating a realistic publication summary for a researcher named \{scientist\_name\} who specializes in molecular design and drug discovery. \\

Generate a list of 5-10 fictional but scientifically plausible publication titles that this researcher might have authored.
Focus on publications relevant to: \{task\_description\} \\

Output format (ONLY titles, one per line, each starting with "- "): \\
- Tetra-substituted pyridinylimidazoles as dual inhibitors of p38$\alpha$ mitogen-activated protein kinase and c-Jun N-terminal kinase 3 for potential treatment of neurodegenerative diseases. \\
- 1,3-Dialkyl-substituted tetrahydropyrimido[1,2-f]purine-2,4-diones as multiple target drugs for the potential treatment of neurodegenerative diseases. \\

Make the publications diverse but coherent with the researcher's expertise in drug discovery and molecular optimization. \\

Output ONLY the list of titles in the exact format shown above.

\end{tcolorbox}
\begin{tcolorbox}[
    colback=gray!5!white,
    colframe=gray!60!black,
    title={
        \parbox[t]{\dimexpr\linewidth-4mm\relax}{%
            \ttfamily
            Prompt for LLM-based profile (molecules) generation
        }
    },
    breakable
]
You are tasked with generating a realistic molecule summary for a researcher named \{scientist\_name\} who specializes in molecular design and drug discovery. \\

Generate a list of 5-10 molecules (as SMILES strings) that this researcher might have worked with in their publications.
Focus on molecules relevant to: \{task\_description\} \\

Output format (SMILES with publication title reference, ending with "...."): \\
- O=C(NCCN1CCOCC1)c1ccnc(-n2ncc3cc(Nc4ccccc4Cl)ccc32)c1 (from: Inhibitors of c-Jun N-terminal kinases: an update....) \\
- CNCC1CCCCC1(OC)c1cccc(OC)c1 (from: In vitro and in vivo evaluation of O-alkyl derivatives of tramadol....) \\

Ensure the SMILES strings are valid and represent drug-like molecules relevant to the task. \\
Output ONLY the list in the exact format shown above.

\end{tcolorbox}

\subsection{Vanilla}

The prompt is employed for vanilla baseline in \cref{subsec:3_3_pmo} and \cref{subsec:3_2_lead_opt}. 

\begin{tcolorbox}[
    colback=gray!5!white,
    colframe=gray!60!black,
    title={
        \parbox[t]{\dimexpr\linewidth-4mm\relax}{%
            \ttfamily
            Prompt for vanilla generation
        }
    },
    breakable
]

You are a researcher specializing in molecular design and drug discovery. \\

Task: \{task\_description\} \\

Generate \{batch\_size\} unique molecules. \\

Output format:
[
    \{"SMILES": "SMILES string"\},
    \{"SMILES": "SMILES string"\},
    ...
] \\

Important:

- Output ONLY the JSON array, no other text or markdown

- Ensure all brackets and quotes are properly closed

- Do NOT wrap in code blocks

- Ensure the SMILES strings are valid and have not proposed in previous proposals

- Do NOT propose the duplicated molecules. Each proposal should be unique.

\end{tcolorbox}

\subsection{Task prompts}

\paragraph{Protein target molecule generation} The task prompt is employed for protein target molecule generation in \cref{subsec:3_1_protein}.

\begin{tcolorbox}[
    colback=gray!5!white,
    colframe=gray!60!black,
    title={
        \parbox[t]{\dimexpr\linewidth-4mm\relax}{%
            \ttfamily
            Task prompt for protein target molecule generation
        }
    },
    breakable
]

Your task is to design a SMILES string for a molecule that maximizes binding affinity to \{protein\_name\}. \\

\# Objective:
Design molecules with high predicted binding affinity (low IC50) to \{protein\_name\}. \\

\# Evaluation Metrics:
- affinity\_pred\_value: log10(IC50) in $\mu$M - LOWER values indicate STRONGER binding \\

\# Protein sequence:
{protein\_sequence} \\

\# Molecular Constraints:
- Must be a valid SMILES string \\

\# Guidelines for High Binding Affinity:

- Include appropriate functional groups for hydrogen bonding (amines, hydroxyls, carbonyls)

- Consider hydrophobic contacts with protein binding pocket (aromatic rings, alkyl chains) 

- Maintain reasonable molecular weight (300-600 Da)

- Include aromatic rings for pi-stacking interactions

- Consider salt bridges with charged residues (carboxylic acids, amines) \\

Generate ONLY the SMILES string with no explanation.

\end{tcolorbox}

\paragraph{Bioactivity-guided molecule generation} The task prompts are employed for bioactivity-guided molecule generation in \cref{subsec:3_3_pmo} and \cref{sec:4_analysis}.

\begin{tcolorbox}[
    colback=gray!5!white,
    colframe=gray!60!black,
    title={
        \parbox[t]{\dimexpr\linewidth-4mm\relax}{%
            \ttfamily
            Task prompt for GSK3$\beta$ bioactivity prediction
        }
    },
    breakable
]

Your task is to design a SMILES string for a molecule that satisfies the following condition: \\

\# Conditions:

- Design a drug-like molecule with high predicted GSK3B (Glycogen Synthase Kinase 3 Beta) inhibitory activity.

- Consider structural features known to enhance kinase binding  affinity. \\

\# IMPORTANT CONSTRAINTS:

- Design drug-like molecules with favorable ADMET properties.

- Maximize the GSK3B binding/inhibitory activity score as high as possible.

- Avoid generating identical structures to provided examples.

- Avoid repeating molecules you already generated.

\end{tcolorbox}
\begin{tcolorbox}[
    colback=gray!5!white,
    colframe=gray!60!black,
    title={
        \parbox[t]{\dimexpr\linewidth-4mm\relax}{%
            \ttfamily
            Task prompt for DRD2 bioactivity prediction

        }
    },
    breakable
]

Your task is to design a SMILES string for a molecule that satisfies the following condition: \\

\# Conditions:

Maximize the probability of binding to the DRD2 receptor (Dopamine Receptor D2). \\

\# IMPORTANT CONSTRAINTS:

- Design drug-like molecules.

- Maximize the DRD2 binding score as high as possible.

- Avoid generating identical structures to provided examples.

- Avoid repeating molecules you already generated.

\end{tcolorbox}

\begin{tcolorbox}[
    colback=gray!5!white,
    colframe=gray!60!black,
    title={
        \parbox[t]{\dimexpr\linewidth-4mm\relax}{%
            \ttfamily
            Task prompt for JNK3 bioactivity prediction
        }
    },
    breakable
]

Your task is to design a SMILES string for a molecule that satisfies the following condition: \\

\# Conditions:

- Design a drug-like molecule with high predicted JNK3 (c-Jun N-terminal kinase 3) inhibitory activity.

- Consider structural features known to enhance kinase inhibition. \\

\# IMPORTANT CONSTRAINTS:

- Design drug-like molecules with favorable ADMET properties.

- Maximize the JNK3 inhibitory activity score as high as possible.

- Avoid generating identical structures to provided examples.

- Avoid repeating molecules you already generated.

\end{tcolorbox}

\paragraph{Goal-directed lead optimization} The task prompt is employed for goal-directed lead optimization in \cref{subsec:3_2_lead_opt}.

\begin{tcolorbox}[
    colback=gray!5!white,
    colframe=gray!60!black,
    title={
        \parbox[t]{\dimexpr\linewidth-4mm\relax}{%
            \ttfamily
            Prompt for goal-directed lead optimization
        }
    },
    breakable
]

Your task is to design a SMILES string for a molecule that optimizes binding affinity to \{protein\_name\}.

You must start from the provided seed molecule, modify it to improve predicted binding affinity (docking score), while strictly satisfying all drug-likeness and similarity constraints. \\

\# Seed Molecule: \{seed\_mol\} \\

\# Hard Constraints (MUST satisfy ALL):

1. Tanimoto Similarity $\geq$ \{sim\_threshold\}

2. QED (drug-likeness) $\geq$ 0.6

3. SA Score (synthetic accessibility) $\leq$ 4

4. NOT identical to seed molecule \\

\# Metric Definitions: \\

\#\# Tanimoto Similarity (target: $\geq$ \{sim\_threshold\})

Measures structural similarity between your molecule and the seed using Morgan fingerprints.

- Score range: 0 (completely different) to 1 (identical)

- Calculated as: (shared features) / (total unique features in both molecules)

- To maintain high similarity: preserve the core scaffold and make only small modifications \\

\#\# QED - Quantitative Estimate of Drug-likeness (target: $\geq$ 0.6)

A composite score combining 8 drug-like properties, ranging 0-1 (higher = more drug-like).

Based on: molecular weight, logP, H-bond donors/acceptors, polar surface area, rotatable bonds, aromatic rings, and structural alerts.

To achieve QED $\geq$ 0.6:

- Molecular Weight: 200-500 Da

- LogP: 0-5

- H-bond donors: $\leq$ 5

- H-bond acceptors: $\leq$ 10

- Rotatable bonds: $\leq$ 10

- Aromatic rings: 1-4 \\

\#\# SA Score - Synthetic Accessibility (target: $\leq$ 4)

Estimates how easy a molecule is to synthesize, ranging 1 (easy) to 10 (very difficult).

Calculated from fragment contributions (based on 1M PubChem molecules) plus complexity penalties for unusual features.

To achieve SA $\leq$ 4:

- Use common, commercially available building blocks

- Avoid: large rings (>8 atoms), bridgehead/spiro atoms, multiple stereocenters

- Prefer simple ring systems (benzene, pyridine, piperidine) \\

\# Recommended Modifications (preserve similarity, maintain drug-likeness): 

- Small substituent changes: -H $\rightarrow$ -F, -CH3 $\rightarrow$ -CF3, -OH $\rightarrow$ -OCH3

- Bioisosteric replacements: benzene $\leftrightarrow$ pyridine, -COOH $\leftrightarrow$ -CONH2

- Methylation of amines: -NH2 $\rightarrow$ -NHCH3

- Small ring modifications: 6-ring $\rightarrow$ 5-ring \\

\# Modifications to AVOID:

- Adding large polycyclic systems (hurts SA)

- Adding > 2 new rings (may hurt similarity)

- Adding unusual functional groups (hurts SA)

- Removing core scaffold elements (hurts similarity) \\

Generate ONLY the SMILES string with no explanation.

\end{tcolorbox}

\section{{Experimental settings}}\label{appx: exp}

We provide the code in \url{https://anonymous.4open.science/r/debate_scientist-AACB}.

\subsection{Hyperparameters}

We configure the collaboration size to $N=50$ scientist agents. In the proposal phase, each agent generates $k=30$ candidate molecules per iteration. To guarantee that the debate yields a sufficient volume of molecular candidates, we set the maximum number of rounds to 20. For all model sampling, we utilize a temperature of 0.7.

\subsection{Computational resource}

We utilized a single NVIDIA RTX A5000 GPU for Boltz-2~\citep{passaro2025boltz} prediction for protein target molecule generation task in \cref{subsec:3_1_protein}.

\newpage
\section{Additional experimental results}\label{appx: exp_add}

\subsection{Detailed examples}

Here, we provide the detailed examples of debate process in \cref{fig:4_1_qualitative}.

\begin{tcolorbox}[
    colback=gray!5!white,
    colframe=gray!60!teal,
    title={
        \parbox[t]{\dimexpr\linewidth-4mm\relax}{%
            \ttfamily
            A detailed example of publication history.
        }
    },
    breakable
]
\{ \\
``title": ``3D-QSAR and docking studies of aminopyridine carboxamide inhibitors of c-Jun N-terminal kinase-1.", \\
``abstract": ``In order to better understand the structural and chemical features of c-Jun N-terminal kinase-1 (JNK-1), which is a member of the mitogen activated protein kinase (MAP kinase) family of enzymes responsible for the serine/threonine phosphorylation of intracellular targets, 3D-QSAR studies of some aminopyridine carboxamides as c-Jun N-terminal kinase inhibitors were performed by comparative molecular field analysis (CoMFA) to rationalize the structural requirements responsible for the inhibitory activity of these compounds. The genetic algorithm of GOLD3.1 has been employed to position 54 aminopyridine carboxamides in the active sites of JNK-1 to determine the probable binding conformation. The docking results provided a reliable conformational alignment scheme for 3D-QSAR model. Based on the docking conformations, highly predictive comparative molecular field analysis (CoMFA) was performed with a cross-validated q(2) of 0.585. The non-cross-validated analysis with six optimum components revealed a conventional r(2) value of 0.988, F=510.200, and an estimated standard error of 0.071. Furthermore, the CoMFA model was mapped back to the binding sites of JNK-1, to get a better understanding of vital interactions between the aminopyridine carboxamides and the kinase. Based on the docking and CoMFA analyses, we have identified some key features in the aminopyridine carboxamides that are responsible for JNK-1 inhibitory activity. The analyses may be used to design more potent aminopyridine carboxamides and predict their activity prior to synthesis." \\
\}, \\
\{ \\
``title": ``Integrating Metabolomics and Proteomics Technologies Provides Insights into the Flavor Precursor Changes at Different Maturity Stages of Arabica Coffee Cherries.", \\
``abstract": ``The metabolic modulation of major flavor precursors during coffee cherry ripening is critical for the characteristic coffee flavor formation. However, the formation mechanism of flavor precursors during coffee cherry ripening remains unknown. In the present study, a colorimeter was employed to distinguish different maturity stages of coffee cherry based on the coffee cherry skin colors, and proteomics and metabolomics profiles were integrated to comprehensively investigate the flavor precursor dynamics involved in Arabica coffee cherry ripening. The data obtained in the present study provide an integral view of the critical pathways involved in flavor precursor changes during coffee cherry ripening. Moreover, the contributions of critical events in regulating the development of flavor precursors during the four ripening stages of coffee cherries, including the biosynthesis and metabolism pathways of organic acids, amino acids, flavonoids, and sugars, are discussed. Overall, a total of 456 difference express metabolites were selected, and they were identified as being concentrated in the four maturity stages of coffee cherries; furthermore, 76 crucial enzymes from the biosynthesis and metabolism of sugars, organic acids, amino acids, and flavonoids contributed to flavor precursor formation. Among these enzymes, 45 difference express proteins that could regulate 40 primary amino acids and organic acids flavor precursors were confirmed. This confirmation indicates that the metabolic pathways of amino acids and organic acids played a significant role in the flavor formation of Arabica coffee cherries during ripening. These results provide new insights into the protease modulation of flavor precursor changes in Arabica coffee cherry ripening." \\
\}, \\

\end{tcolorbox}

\begin{tcolorbox}[
    colback=gray!5!white,
    colframe=gray!60!teal,
    title={
        \parbox[t]{\dimexpr\linewidth-4mm\relax}{%
            \ttfamily
            A detailed example of molecule history.
        }
    },
    breakable
]
\{ \\
``smiles": ``O=C1NC(=O)c2c1c(-c1ccccc1Cl)cc1oc3ccc(O)cc3c21", \\
``inchi\_key": ``KGAPFKLOTLAAMY-UHFFFAOYSA-N", \\
``title": ``3D-QSAR studies of Checkpoint Kinase Weel inhibitors based on molecular docking, CoMFA and CoMSIA.", \\
``pubmed\_id": ``17698256" \\
\}, \\
\{ \\
``smiles": ``CCn1c2ccc(O)cc2c2c3c(c(-c4ccccc4Cl)cc21)C(=O)NC3=O", \\
``inchi\_key": ``PPKWYAYCBJKWOR-UHFFFAOYSA-N", \\
``title": ``3D-QSAR studies of Checkpoint Kinase Weel inhibitors based on molecular docking, CoMFA and CoMSIA.", \\
``pubmed\_id": ``17698256" \\
\}
\end{tcolorbox}

\begin{tcolorbox}[
    colback=gray!5!white,
    colframe=gray!60!teal,
    title={
        \parbox[t]{\dimexpr\linewidth-4mm\relax}{%
            \ttfamily
            A detailed example of proposal phase.
        }
    },
    breakable
]
\{ \\
``SMILES": ``CS(=O)(=O)c1ccc(CNC(=O)c2cc(N)c(C\#N)c(NC3CCCC3)n2)cc1", \\
``rationale": ``This molecule is designed based on structural insights from my paper '3D-QSAR and docking studies of aminopyridine carboxamide inhibitors of c-Jun N-terminal kinase-1.' The core scaffold features an aminopyridine carboxamide, which is a known pharmacophore for JNK inhibition. The sulfonamide group (CS(=O)(=O)) at the para position of the phenyl ring, as seen in several active compounds from that study, is crucial for binding affinity and selectivity. The cyano group (C\#N) adjacent to the pyridine nitrogen enhances electron-withdrawing properties, potentially improving kinase hinge region interaction. The cycloalkylamine substituent (NC3CCCC3) on the pyridine ring is introduced to modulate lipophilicity and metabolic stability, aiming for favorable ADMET properties. This design integrates key features from successful JNK1 inhibitors, adapted here for JNK3, which shares high structural homology in the ATP-binding site." \\
\}, \\
\{ \\
``SMILES": ``CC(=O)Nc1ccc(CNC(=O)c2cc(N)c(C\#N)c(NC3CC3)n2)cc1", \\
``rationale": ``An acetamide group (CC(=O)N) replaces the sulfonamide, exploring alternative hydrogen bond acceptor/donor profiles for JNK3 binding, based on the versatility of amides in kinase inhibitors noted in my QSAR studies. The core cyanoaminopyridine with cyclopropylamine is maintained. The acetyl group may be metabolically labile, offering a potential prodrug effect. This design diversifies from the typical sulfonamide, aiming to discover new interactions while retaining the essential pharmacophore. It reflects the structural variation seen in my work on different triterpene classes, such as cycloartanes versus serratanes, where subtle changes led to distinct biological activities." \\
\}
\end{tcolorbox}

\begin{tcolorbox}[
    colback=gray!5!white,
    colframe=gray!60!teal,
    title={
        \parbox[t]{\dimexpr\linewidth-4mm\relax}{%
            \ttfamily
            A detailed example of critique phase.
        }
    },
    breakable
]
\{ \\
``SMILES": ``CS(=O)(=O)c1ccc(CNC(=O)c2cc(N)c(C\#N)c(NC3CCCC3)n2)cc1", \\
``proposer": ``XXX XXX", \\
``critique": ``Cyclopentylamine is a good hydrophobic group. The sulfonamide remains a concern for metabolism. The molecule is similar to others proposed; consider varying the core scaffold to explore different binding modes, perhaps by incorporating a heterocyclic hinge binder like an indole or purine." \\
\}, \\
\{ \\
``SMILES": ``CS(=O)(=O)c1ccc(CNC(=O)c2cc(N)c(C\#N)c(NC3CCCC3)n2)cc1", \\
``proposer": ``XXX XXX", \\
``critique": ``The cyclobutyl group (CCCC3) is a constrained aliphatic ring, which might improve potency by reducing flexibility. However, the hinge-binding motif is still not optimal. From my research (Papers 1, 6), kinase inhibitors benefit from specific heterocyclic cores." \\
\}
\end{tcolorbox}

\begin{tcolorbox}[
    colback=gray!5!white,
    colframe=gray!60!teal,
    title={
        \parbox[t]{\dimexpr\linewidth-4mm\relax}{%
            \ttfamily
            A detailed example of voting phase.
        }
    },
    breakable
]
\{ \\
``SMILES": ``CS(=O)(=O)c1ccc(CNC(=O)c2cc(N)c(C\#N)c(NC3CCCC3)n2)cc1", \\
``score": 0.9, \\
``justification": ``Based on my 3D-QSAR studies of aminopyridine carboxamide JNK1 inhibitors (Paper 2), this scaffold shows high predicted JNK3 affinity. The sulfonylphenyl linker and cyanoaminopyridine core are key for kinase binding. The cyclobutylamide tail enhances selectivity and ADMET properties, as seen in my previous work." \\
\}, \\ 
\{ \\
``SMILES": ``O=c1ncn2nc(Sc3ccccc3)ccc2c1-c1ccc(C(F)(F)F)cc1", \\
``score": 0.85, \\
``justification": ``This design incorporates a trifluoromethyl group, known to improve lipophilicity and target engagement in kinase inhibitors. My research on antiepileptic drugs (Paper 1,3,6,8,9) highlights the importance of halogenated aromatics for CNS penetration, which is relevant for JNK3 inhibition in neurological contexts. The scaffold is synthetically accessible and avoids toxicity concerns from heavy halogens." \\
\}
\end{tcolorbox}

\subsection{Number of scientists and proposals} 

\begin{figure}[h]
    \centering
    \includegraphics[width=0.5\linewidth]{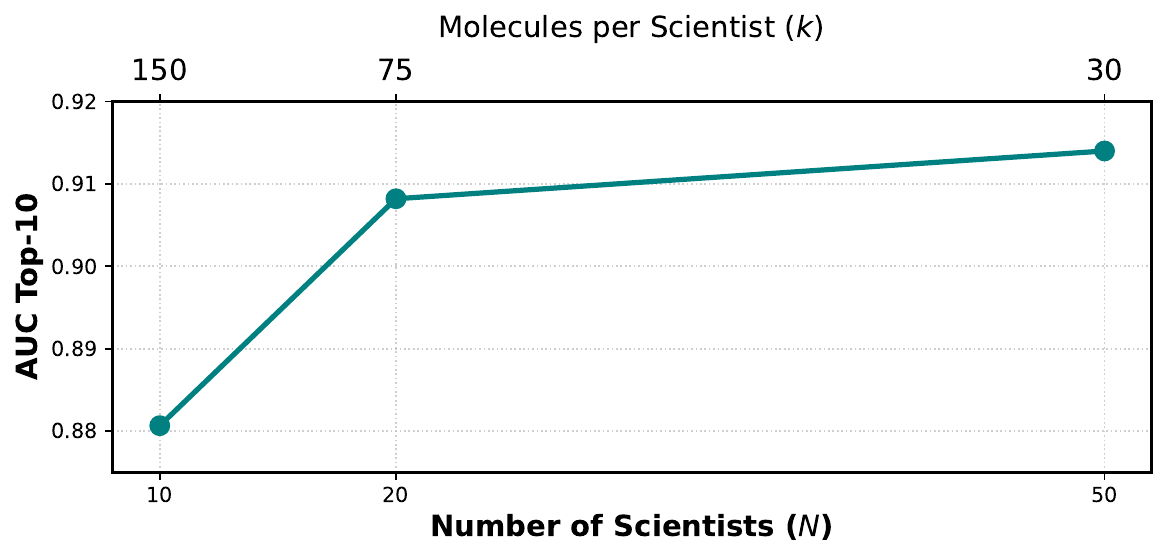}
    \caption{\textbf{Fixed total number of proposals.}}
    \label{fig:placeholder}
\end{figure}

To analyze the impact of the number of scientists and the number of proposals per scientist, we evaluate the interplay between the number of scientists that debate ($N$) and the number of proposals that one scientist proposes per round ($k$) in JNK bioactivity optimization task.

First, we fix the total number of proposals $N\times k=1,500$. This allows us to investigate the trade-off between scientist diversity (increasing $N$) against exploration depth (increasing $k$). By varying the composition, we observe whether the scientist diversity is more critical than the depth of expertise.

We provide the results in \cref{fig:3_4_ablation_num}. This demonstrates demonstrate that increasing the number of scientists ($N$) consistently enhances discovery performance across both experimental settings. In the fixed total budget scenario, the system maintains relatively high AUC scores even at low N, suggesting that increased exploration depth ($k$) can partially compensate for limited expertise diversity. 

\subsection{Detailed protein-conditioned molecule generation results}

\begin{table*}[h]
\centering
\caption{\textbf{Results of protein target molecule generation (binding affinity).} \textbf{Bold} highlights the best scores.}
\label{tab:3_1_protein}
\resizebox{0.98\linewidth}{!}{
\begin{tabular}{c cccccccc cccccccc}
\toprule[1.25pt]
& \multicolumn{2}{c}{\textbf{TYK2}} &  \multicolumn{2}{c}{\textbf{JNK1}} & \multicolumn{2}{c}{\textbf{CDK2}} & \multicolumn{2}{c}{\textbf{P38}} &  \multicolumn{2}{c}{\textbf{CA2}} & \multicolumn{2}{c}{\textbf{DHFR}} & \multicolumn{2}{c}{\textbf{FABP4}} & \multicolumn{2}{c}{\textbf{THROMBIN}} \\
\cmidrule(lr){2-3} \cmidrule(lr){4-5} \cmidrule(lr){6-7} \cmidrule(lr){8-9} \cmidrule(lr){10-11} \cmidrule(lr){12-13} \cmidrule(lr){14-15} \cmidrule(lr){16-17}

& \textbf{Top1} & \textbf{Top10} & \textbf{Top1} & \textbf{Top10} & \textbf{Top1} & \textbf{Top10} & \textbf{Top1} & \textbf{Top10} & \textbf{Top1} & \textbf{Top10} & \textbf{Top1} & \textbf{Top10} & \textbf{Top1} & \textbf{Top10} & \textbf{Top1} & \textbf{Top10} \\
\midrule
\textsc{VanillaDebate} & \phantom{0}-9.92 & \phantom{0}-9.48 & \phantom{0}-8.99 & \phantom{0}-8.75 & -10.00 & \phantom{0}-9.20 & \phantom{0}-9.42 & \phantom{0}-9.04 & -11.09 & -10.92 & \phantom{0}-8.68 & \phantom{0}-8.17 & \phantom{0}-8.40 & \phantom{0}-7.87&\phantom{0}-9.35 & \phantom{0}-8.70 \\
\textsc{KeywordDebate} & \phantom{0}-9.73 & \phantom{0}-9.40 & \phantom{0}-9.49 & \phantom{0}-9.13 & \phantom{0}-9.48 & \phantom{0}-9.03 & \phantom{0}-9.77 & \phantom{0}-9.08 & -11.00 & -10.62 & \phantom{0}-9.22 & \phantom{0}-8.86 & \phantom{0}-9.00 & \phantom{0}-8.13 & \phantom{0}-9.73 & \phantom{0}-8.50 \\

\midrule
\methodname (Ours)& \textbf{-11.97} & \textbf{-10.71} & \textbf{-10.51} & \textbf{-10.08} & \textbf{-10.52} & \textbf{-10.12} & \textbf{-10.88} & \textbf{-10.37} & \textbf{-12.76} & \textbf{-11.68} & \textbf{-11.36} & \textbf{-10.12} & \textbf{\phantom{0}-9.65} & \textbf{\phantom{0}-9.23} & \textbf{-11.33} & \textbf{-10.71} \\
\bottomrule[1.25pt]
\end{tabular}}
\end{table*}

\begin{table*}[t]
\centering
\caption{\textbf{Results of protein target molecule generation (diversity).} \textbf{Bold} highlights the best scores.}
\label{tab:3_1_protein_diversity}
\resizebox{0.98\linewidth}{!}{
\begin{tabular}{c cccccccc cccc}
\toprule[1.25pt]
& \multicolumn{3}{c}{\textbf{TYK2}} &  \multicolumn{3}{c}{\textbf{JNK1}} & \multicolumn{3}{c}{\textbf{CDK2}} & \multicolumn{3}{c}{\textbf{P38}} \\
\cmidrule(lr){2-4} \cmidrule(lr){5-7} \cmidrule(lr){8-10} \cmidrule(lr){11-13} 
& \textbf{IntDiv} & \textbf{$\#\text{Cir.}_{0.75}$} & \textbf{$\#\text{Cir.}_{0.85}$} & \textbf{IntDiv} & \textbf{$\#\text{Cir.}_{0.75}$} & \textbf{$\#\text{Cir.}_{0.85}$}& \textbf{IntDiv} & \textbf{$\#\text{Cir.}_{0.75}$} & \textbf{$\#\text{Cir.}_{0.85}$}& \textbf{IntDiv} & \textbf{$\#\text{Cir.}_{0.75}$} & \textbf{$\#\text{Cir.}_{0.85}$}\\
\midrule
\textsc{VanillaDebate} & 0.8079 & 53 & 9 & 0.7957 & 40 & 7 & 0.8021 & 47 & 9 & 0.8085 & 55 & 9 \\
\textsc{KeywordDebate} & 0.7941 & 42 & 7 & 0.8103 & 49 & 8 & 0.8034 & 41 & 6 & 0.8096 & 45 & 8 \\
\midrule
\methodname (Ours)& \textbf{0.8695} & \textbf{203} & \textbf{35} & \textbf{0.8691} & \textbf{229} & \textbf{38} & \textbf{0.8661} & \textbf{211} & \textbf{45} & \textbf{0.8699} & \textbf{239} & \textbf{43}\\
\bottomrule[1.25pt]
\end{tabular}}

\resizebox{0.98\linewidth}{!}{
\begin{tabular}{c cccccccc cccc}
\toprule[1.25pt]
& \multicolumn{3}{c}{\textbf{CA2}} &  \multicolumn{3}{c}{\textbf{DHFR}} & \multicolumn{3}{c}{\textbf{FABP4}} & \multicolumn{3}{c}{\textbf{THROMBIN}} \\
\cmidrule(lr){2-4} \cmidrule(lr){5-7} \cmidrule(lr){8-10} \cmidrule(lr){11-13} 
& \textbf{IntDiv} & \textbf{$\#\text{Cir.}_{0.75}$} & \textbf{$\#\text{Cir.}_{0.85}$} & \textbf{IntDiv} & \textbf{$\#\text{Cir.}_{0.75}$} & \textbf{$\#\text{Cir.}_{0.85}$}& \textbf{IntDiv} & \textbf{$\#\text{Cir.}_{0.75}$} & \textbf{$\#\text{Cir.}_{0.85}$}& \textbf{IntDiv} & \textbf{$\#\text{Cir.}_{0.75}$} & \textbf{$\#\text{Cir.}_{0.85}$}\\
\midrule
\textsc{VanillaDebate} & 0.7468 & 20 & 6 & 0.7821 & 37 & 9 & 0.7529 & 35 & 7 & 0.7408 & 27 & 5  \\
\textsc{KeywordDebate} & 0.7274 & 23 & 6 & 0.7990 & 42 & 8 & 0.7130 & 27 & 7 & 0.7539 & 26 & 5 \\

\midrule
\methodname (Ours)& \textbf{0.8655} & \textbf{139} & \textbf{46} & \textbf{0.8679} & \textbf{192} & \textbf{40} & \textbf{0.8714} & \textbf{250} & \textbf{44} & \textbf{0.8576} & \textbf{165} & \textbf{36} \\

\bottomrule[1.25pt]
\end{tabular}}
\end{table*}

Here, we provide detailed protein conditioned molecule generation results in \cref{tab:3_1_protein}.

\subsection{Additional PMO tasks}

\begin{table*}[h]
    \centering
    \caption{\textbf{Results of PMO-1K benchmark.} Tasks are assessed using top-10 AUC. We mark the best result in \textbf{bold} and \textcolor{teal}{teal} highlights the improvement to the vanilla-debate.}
    \label{tab_appx:pmo}
    \resizebox{\linewidth}{!}{
    \begin{tabular}{c ccc cccccccc ccc }
    \toprule[1.25pt] &
      \multicolumn{3}{c}{\textbf{Bioactivity}} & \multicolumn{7}{c}{\textbf{Multi property optimization}} & \multicolumn{3}{c}{\textbf{Rediscovery}} \\
    \cmidrule(lr){2-4} \cmidrule(lr){5-11} \cmidrule(lr){12-14}
    \textbf{Model} & \textbf{GSK3$\beta$} & \textbf{DRD2} & \textbf{JNK3}  & \textbf{Amlo.} & \textbf{Fexo.} & \textbf{Osim.} & \textbf{Peri.} & \textbf{Rano.} & \textbf{Sita.} & \textbf{Zale.} & \textbf{Cele.} & \textbf{Thio.} & \textbf{Trog.}  \\
    \midrule
    GP BO        & 0.611 & 0.857 &0.346 & 0.519 & 0.707 & 0.766 & 0.458 & 0.701 & 0.232 & 0.392 & 0.411 & 0.351 & 0.313  \\
    REINVENT    & 0.589 &  0.775& 0.315   & 0.472 & 0.650 & 0.737 & 0.404 & 0.574 & 0.261 & 0.406 & 0.370 & 0.311 & 0.246  \\
    LICO-L   & 0.617 &0.859 &0.336     & 0.541 & 0.700 & 0.759 & 0.473 & 0.687 & \textbf{0.315} & 0.404 & 0.447 & 0.343 & 0.292  \\
    Genetic GFN  &0.637 &0.809&0.409  & 0.534 & 0.682 & 0.763 & 0.462 & 0.623 & 0.227 & 0.400 & 0.447 & 0.377 & 0.277  \\
    Graph GA    &  0.523 &0.833&0.301  & 0.501 & 0.666 & 0.751 & 0.435 & 0.620 & 0.229 & 0.374 & 0.424 & 0.322 & 0.267  \\
    Aug. Mem.   &  0.539 &0.795&0.294   & 0.489 & 0.679 & 0.761 & 0.422 & 0.614 & 0.245 & 0.415 & 0.385 & 0.336 & 0.262  \\
    MOLLEO-B    &  0.397& 0.910&0.186 & 0.637 & 0.674 & 0.779 & 0.655 & 0.640 & 0.193 & 0.392 & 0.402 & 0.416 & 0.302  \\
    MOLLEO-D     &   0.496 & 0.812&0.342  & 0.540 & 0.680 & 0.753 & 0.422 & 0.516 & 0.328 & 0.409 & 0.512 & 0.478 & 0.387  \\
    MT-Mol     &  0.308 &0.756&0.125  & 0.647 & 0.883 & 0.796 & 0.542 & 0.233 & 0.067 & 0.625 & 0.867 & 0.719 & \textbf{0.841}  \\
    \midrule
    Vanilla & 0.419 & 0.921 & 0.310 & 0.854 & 0.530 & 0.877 & 0.678 & 0.642 & 0.099 & 0.708 & 0.825 & \textbf{0.845} & 0.825  \\
    
    Vanilla-debate &  0.477 & 0.902 & 0.161  & \textbf{0.856}   & \textbf{0.935}   & 0.939   & 0.769   & 0.636   & 0.310   & \textbf{0.740}   & 0.819   & 0.808   & 0.820      \\
    \midrule
    \methodname (Ours)& \textbf{\textcolor{teal}{0.942}} & \textbf{\textcolor{teal}{0.950}} & \textbf{\textcolor{teal}{0.914}} &{0.845} & {0.925} & \textbf{\textcolor{teal}{0.941}} & \textbf{\textcolor{teal}{0.775}} & \textbf{\textcolor{teal}{0.848}} & 0.225 & 0.730 & \textbf{\textcolor{teal}{0.821}} & {\textcolor{teal}{0.831}} & \textcolor{teal}{0.838}  \\
    \bottomrule[1.25pt]
    \end{tabular}
    }
\end{table*}

For completeness, we provide additional results on the PMO benchmark~\citep{gao2022sample}, including multi-property optimization and molecule rediscovery tasks. Although these tasks were excluded from the main text due to their nature as arithmetic structural puzzles, we report their performance in \cref{tab_appx:pmo}. We observe that \methodname still achieves consistent performance improvements in most cases compared to the vanilla debate. However, we emphasize that these metrics are less indicative of our framework's true utility, as they do not require the individual profile-grounded reasoning or the broad chemical space exploration that \methodname is designed to facilitate.


\end{document}